\UseRawInputEncoding
\documentclass[10pt,twocolumn]{article} 
\pdfoutput=1
\usepackage{simpleConference}
\usepackage{times}
\usepackage{graphicx}
\usepackage{amsmath}
\usepackage{amssymb}
\usepackage{booktabs}
\usepackage{algorithm}
\usepackage{algorithmic}
\usepackage{makecell}
\usepackage{cuted}
\usepackage{float}
\usepackage{capt-of}
\usepackage{hyperref}
\hypersetup{pagebackref,breaklinks,colorlinks}

\begin{document}

\title{Few-shot Image Generation with Diffusion Models}

\author{JingYuan Zhu \\
Tsinghua University, China \\
jy-zhu20@mails.tsinghua.edu.cn\\
\and
Huimin Ma \\
University of Science and Technology Beijing, China \\
mhmpub@ustb.edu.cn  \\
\and
Jiansheng Chen \\
University of Science and Technology Beijing, China \\
jschen@ustb.edu.cn  \\
\and
Jian Yuan  \\
Tsinghua University, China \\
jyuan@tsinghua.edu.cn  \\
}

\maketitle
\thispagestyle{empty}

\begin{abstract}
   Denoising diffusion probabilistic models (DDPMs) have been proven capable of synthesizing high-quality images with remarkable diversity when trained on large amounts of data. However, to our knowledge, few-shot image generation tasks have yet to be studied with DDPM-based approaches. Modern approaches are mainly built on Generative Adversarial Networks (GANs) and adapt models pre-trained on large source domains to target domains using a few available samples. In this paper, we make the first attempt to study when do DDPMs overfit and suffer severe diversity degradation as training data become scarce. Then we fine-tune DDPMs pre-trained on large source domains to solve the overfitting problem when training data is limited. Although the directly fine-tuned models accelerate convergence and improve generation quality and diversity compared with training from scratch, they still fail to retain some diverse features and can only produce coarse images. Therefore, we design a DDPM pairwise adaptation (DDPM-PA) approach to optimize few-shot DDPM domain adaptation. DDPM-PA efficiently preserves information learned from source domains by keeping the relative pairwise distances between generated samples during adaptation. Besides, DDPM-PA enhances the learning of high-frequency details from source models and limited training data. DDPM-PA further improves generation quality and diversity and achieves results better than current state-of-the-art GAN-based approaches. We demonstrate the effectiveness of our approach on a series of few-shot image generation tasks qualitatively and quantitatively. 
\end{abstract}

\section{Introduction}
\label{sec:intro}
 Recent advances in generative models including GANs \cite{NIPS2014_5ca3e9b1, DBLP:conf/iclr/BrockDS19, Karras_2019_CVPR, Karras_2020_CVPR, Karras2021}, variational autoencoders (VAEs) \cite{kingma2013auto, rezende2014stochastic, vahdat2020nvae}, and autoregressive models \cite{van2016conditional,chen2018pixelsnail,henighan2020scaling} have realized high-quality image generation with great diversity. Diffusion probabilistic models \cite{sohl2015deep} are introduced to match data distributions by learning to reverse multi-step noising processes. Ho et al. \cite{NEURIPS2020_4c5bcfec} demonstrate the capability of DDPMs to produce high-quality results. Following works \cite{song2020improved, dhariwal2021diffusion, nichol2021improved,kingma2021variational} further optimize the noise addition schedules, network architectures, and optimization targets of DDPMs. Besides, classifier guidance is added to realize DDPM-based conditional image generation \cite{dhariwal2021diffusion}. DDPMs have shown excellent generation results competitive with GANs \cite{Karras_2020_CVPR, DBLP:conf/iclr/BrockDS19} on datasets including CIFAR-10 \cite{krizhevsky2009learning}, LSUN \cite{yu2015lsun}, and ImageNet \cite{van2016conditional}. Moreover, DDPMs have also achieved compelling results in generating videos \cite{ho2022video, harvey2022flexible, yang2022diffusion, zhang2022motiondiffuse}, audios \cite{kong2020diffwave, austin2021structured}, point clouds \cite{zhou20213d, luo2021diffusion, lyu2021conditional, liu2022let}, and biological structures \cite{xu2022geodiff,hoogeboom2022equivariant,geossl}.  

Modern DDPMs depend on large amounts of data to train the millions of parameters in their networks like other generative models, which tend to overfit seriously and fail to produce high-quality images with considerable diversity when training data is limited. Unfortunately, it is not always possible to obtain abundant data under some circumstances. A series of GAN-based approaches \cite{wang2018transferring, ada,mo2020freeze, wang2020minegan, ewc, ojha2021few-shot-gan, zhao2022closer} have been proposed to adapt models pre-trained on large-scale source datasets to target datasets using a few available training samples (e.g., 10 images). These approaches utilize knowledge from source models to relieve overfitting but can only achieve limited quality and diversity. In addition, the performance of DDPMs trained on limited data and practical DDPM-based few-shot image generation approaches remain to be investigated.

We first evaluate the performance of DDPMs trained on small-scale datasets and show that DDPMs suffer similar overfitting problems to other modern generative models when trained on limited data. Then we fine-tune pre-trained DDPMs on target domains using limited data directly. The fine-tuned DDPMs achieve faster convergence and diverse generated samples compared with DDPMs trained from scratch but still get results sharing similar features and missing high-frequency details. To this end, we introduce the DDPM-PA approach, which keeps the relative distance between generated samples and realizes high-frequency details enhancement during domain adaptation to achieve high-quality few-shot image generation with great diversity. 

To sum up, the contributions of our work are:

\begin{itemize}
    \item We make the first attempt to study when do DDPMs overfit as training data become scarce and design a Nearest-LPIPS metric to evaluate generation diversity. 
    \item We propose a pairwise similarity loss to keep the relative pairwise distances between generated samples during DDPM domain adaptation for greater diversity. 
    \item We design a high-frequency details enhancement approach from two perspectives, including preserving details provided by source models and learning more details from limited data during DDPM domain adaptation for finer quality.
    \item  We demonstrate the effectiveness of DDPM-PA qualitatively and quantitatively on a series of few-shot image generation tasks and show that DDPM-PA achieves better generation quality and diversity than current state-of-the-art GAN-based approaches.
    
\end{itemize}

\section{Related Work}
\subsection{Diffusion Denoising Probabilistic Models}
\textbf{DDPMs Formulation} Given training images $x_0$ following the distribution $q(x_0)$, DDPMs define a forward noising (diffusion) process $q$ adding Gaussian noises with variance $\beta_t \in (0,1)$ at diffusion step $t$ and produce the noised image sequences: $x_1,x _2,...,x_T$ as follows:
\begin{align}
\label{eq1}
    q(x_1,x_2,...,x_T|x_0) &:= \prod_{t=1}^{T} q(x_t|x_{t-1}), \\
\label{eq2}
    q(x_t|x_{t-1}) &:= \mathcal{N}(x_t;\sqrt{1-\beta_t}x_{t-1},\beta_t \mathbf{I}).
\end{align}
We can sample an arbitrary step of the noising process conditioned on $x_0$ as follows:
\begin{align}
\label{eq3}
        q(x_t|x_0) &= \mathcal{N}(x_t;\sqrt{\overline{\alpha}_t}x_0,(1-\overline{\alpha}_t)\mathbf{I}), \\
        x_t &= \sqrt{\overline{\alpha}_t}x_0 + \sqrt{1-\overline{\alpha}_t}\epsilon, \label{eq4}
\end{align}
where $\alpha_t:=1-\beta_t$, $\overline{\alpha}_t:=\prod_{s=0}^{t}\alpha_s$, and $\epsilon$ represents Gaussian distributions following $\mathcal{N}(0,\mathbf{I})$. Based on the Bayes theorem, we have the posterior $q(x_{t-1}|x_t,x_0)$:
\begin{equation}
    q(x_{t-1}|x_t,x_0)=\mathcal{N}(x_{t-1};\hat{\mu}_t(x_t,x_0),\hat{\beta}_t\mathbf{I}),
\end{equation}
where $\hat{\mu}_t(x_t,x_0)$ and $\hat{\beta}_t$ are defined in terms of $x_0, x_t$ and the variance as follows:
\begin{align}
    \hat{\mu}_t(x_t,x_0)&:=\frac{\sqrt{\overline{\alpha}_{t-1}}\beta_t}{1-\overline{\alpha}_t}x_0 + \frac{\sqrt{\alpha_t}(1-\overline{\alpha}_{t-1})}{1-\overline{\alpha}_t}x_t, \label{eq6}\\ 
    \hat{\beta}_t&:=\frac{1-\overline{\alpha}_{t-1}}{1-\overline{{\alpha}}_t}\beta_t.
\end{align}

With a large enough $T$, the noised image $x_T$ almost follows an isotropic Gaussian distribution. Therefore, we can randomly sample a $x_T$ from $\mathcal{N}(0,\mathbf{I})$ and apply the reverse distribution $q(x_{t-1}|x_t)$ to reverse the diffusion process and get the sample following $q(x_0)$. DDPMs employ a UNet-based neural network to approximate the reverse distribution $q(x_{t-1}|x_t)$ as:
\begin{align}
\label{eq8}
    p_{\theta}(x_{t-1}|x_t):=\mathcal{N}(x_{t-1};\mu_{\theta}(x_t,t),\Sigma_{\theta}(x_t,t)).
\end{align}
Naturally, we can train the network to predict $x_0$ and apply it to Equation \ref{eq6} to parameterize the reverse distribution mean $\mu_{\theta}(x_t,t)$. Besides, we can also derive the prediction of $\mu_{\theta}(x_t,t)$ combing Equations \ref{eq4} and \ref{eq6} as:
\begin{align}
    \mu_{\theta}(x_t,t)=\frac{1}{\sqrt{\alpha_t}}(x_t-\frac{\beta_t}{\sqrt{1-\overline{\alpha}_t}}\epsilon_{\theta}(x_t,t)),
\end{align}
 if we train the network as a function approximator $\epsilon_{\theta}(x_t,t)$ to predict the noise $\epsilon$ in Equation \ref{eq4}. Ho et al. \cite{NEURIPS2020_4c5bcfec} demonstrate that predicting $\epsilon$ performs well and achieves high-quality results using a reweighted loss function:
\begin{align}
\label{loss_simple}
    \mathcal{L}_{simple}=E_{t,x_0,\epsilon}\left[||\epsilon-\epsilon_{\theta}(x_t,t)||\right]^2.
\end{align}
In Ho et al.'s work \cite{NEURIPS2020_4c5bcfec}, the variance $\Sigma_{\theta}(x_t,t)$ is fixed as a constant $\sigma_t^2 \mathbf{I}$, where $\sigma_t^2=\beta_t$ and is not learned. The network is only trained to learn the model mean $\mu_{\theta}(x_t,t)$ through predicting noises with $\epsilon_\theta(x_t,t)$. Following works \cite{nichol2021improved, kingma2021variational} propose to optimize the variational lower bound (VLB) and guide the learning of $\Sigma_{\theta}(x_t,t)$ with an additional optimization term $L_{vlb}$ as follows:
\begin{align}
    L_{vlb}:&=L_0+L_1+...+L_{T-1}+L_T, \\
    L_0:&=-log\ p_{\theta}(x_0|x_1), \\
    L_{t-1}:&=D_{KL}(q(x_{t-1}|x_t,x_0)\,||\,p_{\theta}(x_{t-1}|x_t)), \\
    L_T:&=D_{KL}(q(x_T|x_0)\,||\,p(x_T)),
\end{align}
where $D_{KL}$ represents the Kullback-Leibler (KL) divergence used to evaluate the distance between distributions.

\textbf{Fast Sampling} DDPMs need time-consuming iterative processes to realize sampling following Equation \ref{eq8}. Recent works including DDIM and gDDIM \cite{song2020denoising, zhang2022gddim} extend the original DDPMs to non-Markovian cases for fast sampling. DPM-solver \cite{lu2022dpm,lu2022dpm2} presents a theoretical formulation for the solution of probability flow ordinary differential equations (ODEs) and achieves a fast ODE solver needing only 10 steps for DDPM sampling. Diffusion Exponential Integrator Sampler (DEIS) \cite{zhang2022fast} makes use of an exponential integrator to approximately calculate the solution of ODEs. Karras et al. \cite{karras2022elucidating} present a design space to identify changes in both the sampling and training processes and achieve state-of-the-art FID \cite{heusel2017gans} on CIFAR-10 \cite{krizhevsky2009learning} under a class-conditional setting.

\textbf{Applications} DDPMs have already been applied to many aspects of applications such as image super-resolution \cite{li2022srdiff, saharia2022image, rombach2022high,ho2022cascaded}, image translation \cite{saharia2022palette, ozbey2022unsupervised}, semantic segmentation \cite{baranchuk2021label, asiedu2022decoder}, few-shot generation for unseen classes \cite{giannone2022few,sinha2021d2c}, and natural language processing \cite{austin2021structured, li2022diffusion, chen2022analog}. Besides, DDPMs are combined with other generative models including GANs \cite{xiao2021tackling, wang2022diffusion}, VAEs \cite{vahdat2021score, huang2021variational, luo2022understanding}, and autoregressive models \cite{rasul2021autoregressive, hoogeboom2021autoregressive}. Different from existing works, this paper focuses on model-level, unconditional, few-shot image generation with DDPM-based approaches.

\subsection{Few-shot Image Generation}
    Few-shot image generation aims to achieve high-quality generation with great diversity using only a few available training samples. However, modern generative models easily overfit and suffer severe diversity degradation when trained on limited data (e.g., 10 images). They tend to replicate training samples instead of generating diverse images following similar distributions. GAN-based few-shot image generation approaches mainly follow TGAN \cite{wang2018transferring} to adapt GANs pre-trained on large source domains, including ImageNet \cite{van2016conditional}, LSUN \cite{yu2015lsun}, and FFHQ \cite{Karras_2020_CVPR}, to related target domains with limited data. Augmentation approaches \cite{tran2021data, zhao2020differentiable, zhao2020image, ada} also help improve generation diversity. BSA \cite{noguchi2019image} fixes all the parameters except for the scale and shift parameters in the generator. FreezeD \cite{mo2020freeze} freezes parameters in high-resolution layers of the discriminator to relieve overfitting. MineGAN \cite{wang2020minegan} uses additional fully connected networks to modify noise inputs for the generator. EWC \cite{ewc} makes use of elastic weight consolidation to regularize the generator by making it harder to change the critical weights which have higher Fisher information values. CDC \cite{ojha2021few-shot-gan} proposes a cross-domain consistency loss and patch-level discrimination to build a correspondence between source and target domains. DCL \cite{zhao2022closer} utilizes contrastive learning to push away the generated samples from real images and maximize the similarity between corresponding image pairs in source and target domains. The proposed DDPM-PA approach follows similar strategies to adapt models pre-trained on large source domains to target domains. Our experiments show that DDPMs are qualified for few-shot image generation tasks and can achieve better results than current state-of-the-art GAN-based approaches in generation quality and diversity.

\section{DDPMs Trained on Small-scale Datasets}
\label{section3}
To evaluate the performance of DDPMs when training data is limited, we train DDPMs on small-scale datasets containing various numbers of images from scratch. We analyze generation diversity qualitatively and quantitatively to study when do DDPMs overfit as training samples decrease. 

\textbf{Basic Setups} We sample 10, 100, and 1000 images from FFHQ-babies (Babies), FFHQ-sunglasses (Sunglasses) \cite{ojha2021few-shot-gan}, and LSUN Church \cite{yu2015lsun} respectively as small-scale training datasets. The image resolution of all the datasets is set as $256\times 256$. We follow the model setups in prior works \cite{nichol2021improved, dhariwal2021diffusion} used for LSUN $256^2$ \cite{yu2015lsun}. The max diffusion step $T$ is set as 1000. We use a learning rate of 1e-4 and a batch size of 48. We train DDPMs for $40K$ iterations on datasets containing 10 or 100 images and $60K$ iterations on datasets containing 1000 images empirically.

\begin{figure*}[t]
    \centering
    \includegraphics[width=1.0\linewidth]{ 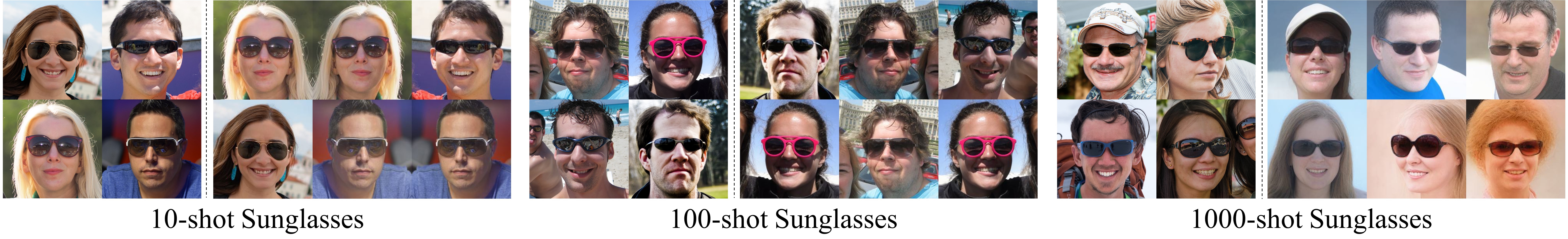}
    \caption{ For small-scale Sunglasses datasets containing 10, 100, and 1000 images, \textbf{Left}: samples picked from the small-scale datasets, \textbf{Right}: samples produced by DDPMs trained on the small-scale datasets from scratch.}
    \label{scratch}
\end{figure*}

\textbf{Qualitative Evaluation} In Fig. \ref{scratch}, we visualize the generated samples of DDPMs trained from scratch on few-shot Sunglasses datasets and provide some training samples for comparison (more generated and training samples are added in Appendix \ref{appendix_scratch}). We observe that DDPMs overfit and tend to replicate training samples when datasets are limited to 10 or 100 images. Since some training samples are flipped in the training process as a step of data augmentation, we can also find some generated images symmetric to the training samples. While for datasets containing 1000 images, DDPMs can generate diverse samples following similar distributions of training samples instead of replicating them. The overfitting problem is relatively alleviated. However, the generated samples are coarse and lack high-frequency details compared with training samples.

\textbf{Quantitative Evaluation} LPIPS \cite{zhang2018unreasonable} is proposed to evaluate the perceptual distances \cite{johnson2016perceptual} between images. We propose a Nearest-LPIPS metric based on LPIPS to evaluate the generation diversity of DDPMs trained on small-scale datasets. More specifically, we first generate 1000 images randomly and find the most similar training sample having the lowest LPIPS distance to each generated sample. Nearest-LPIPS is defined as the LPIPS distances between generated samples and the most similar training samples in correspondence, averaged over all the generated samples. If a generative model reproduces the training samples exactly, the Nearest-LPIPS metric will have a score of zero. Larger Nearest-LPIPS values indicate lower replication rates and greater diversity relative to training samples. 

\begin{figure*}[t]
    \centering
    \includegraphics[width=1.0\linewidth]{ 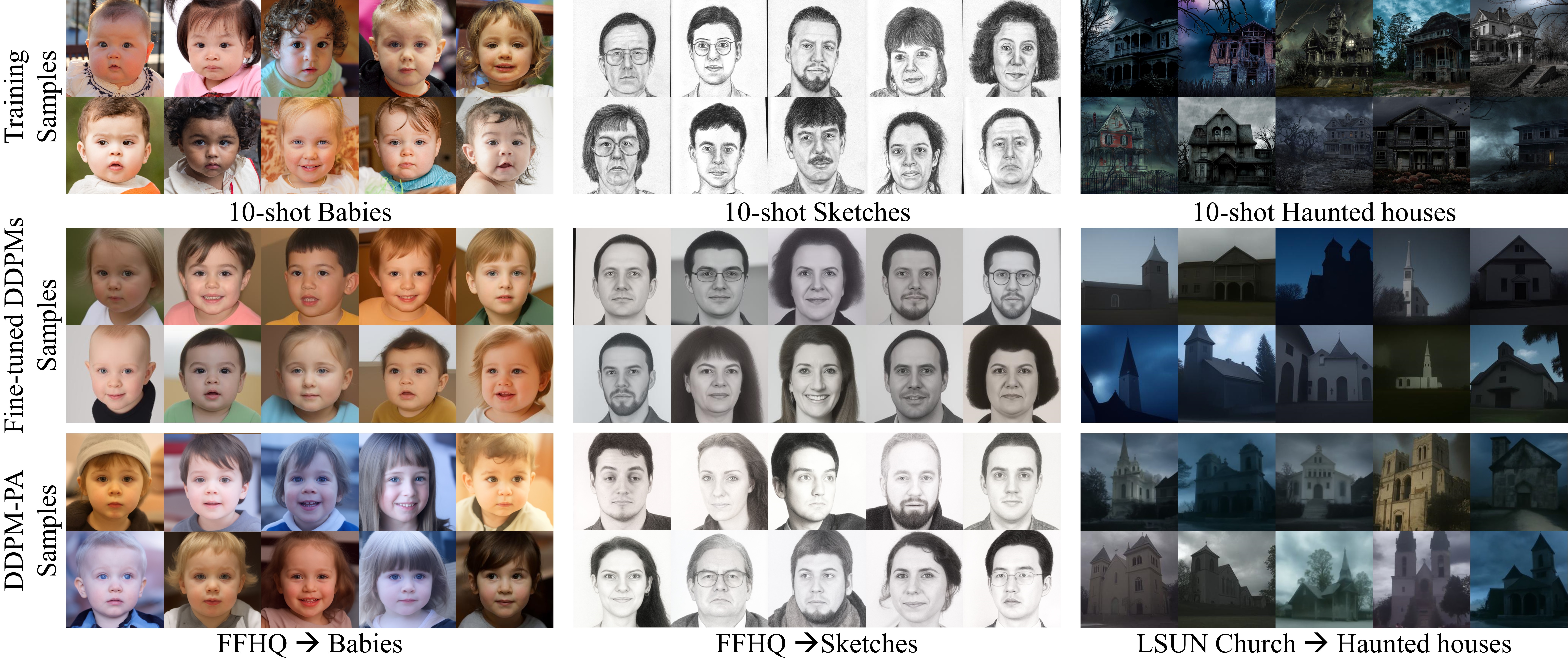}
    \caption{DDPM-based image generation samples on 10-shot FFHQ $\rightarrow$ Babies, FFHQ $\rightarrow$ Sketches, and LSUN Church $\rightarrow$ Haunted houses.}
    \label{result1}
\end{figure*}

\begin{table}[tbp]
\centering
\begin{tabular}{c|c|c|c}
Number of Samples & Babies & Sunglasses & Church \\
\hline
$10$ & $0.2875$ & $0.3030$ & $0.3136$ \\
$100$ & $0.3152$ & $0.3310$ & $0.3327$ \\
$1000$  & $\pmb{0.4658}$ & $\pmb{0.4819}$ & $\pmb{0.5707}$ \\
\hline
$10$ (+ flip) & $0.1206$ & $0.1217$ & $0.0445$\\
$100$ (+ flip) & $0.1556$ & $0.1297$ & $0.1177$ \\
$1000$ (+ flip) & $\pmb{0.4611}$ & $\pmb{0.4726}$ & $\pmb{0.5625}$ \\
\end{tabular}
\caption{Nearest-LPIPS ($\uparrow$) results of DDPMs trained from scratch on several small-scale datasets.}
\label{nearlpips}
\end{table}

\begin{figure}[t]
    \centering
    \includegraphics[width=1.0\linewidth]{ 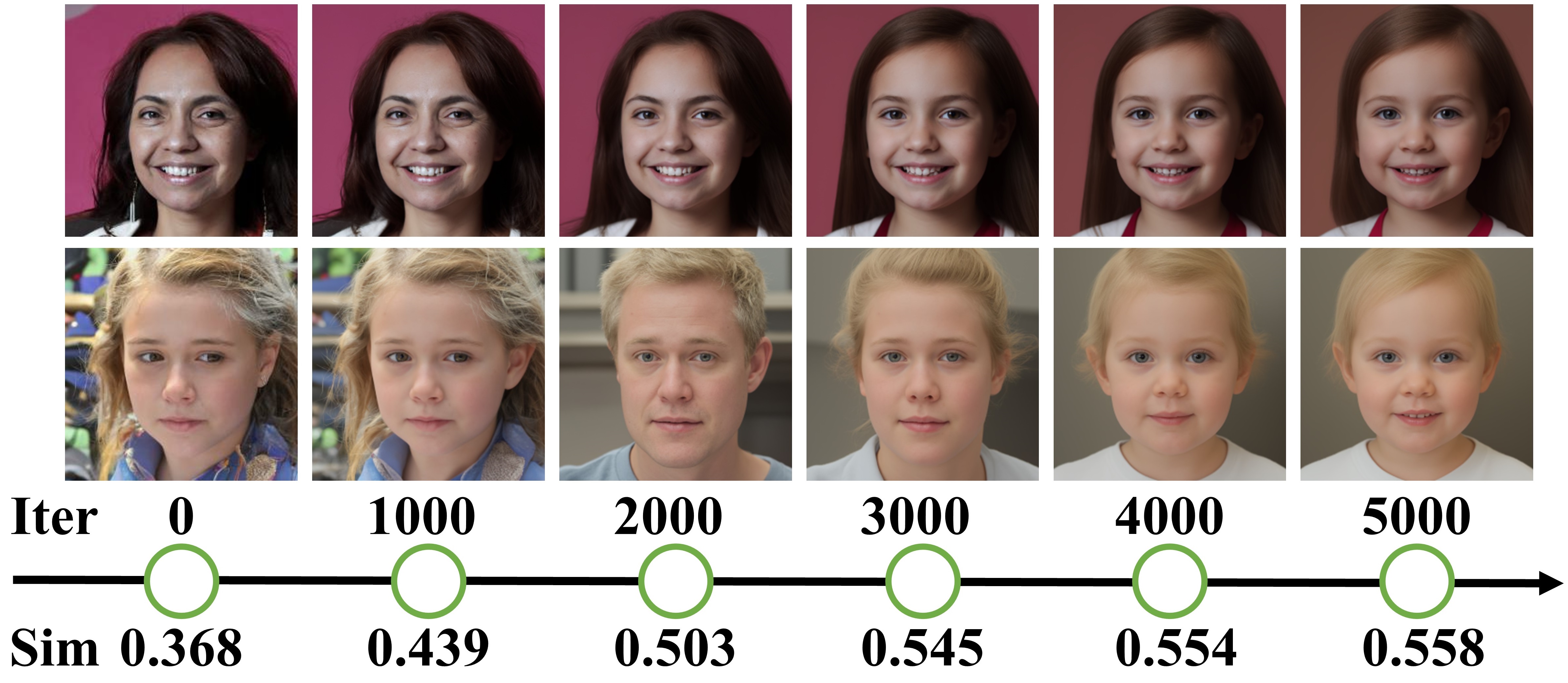}
    \caption{Two samples synthesized from fixed noise inputs by the directly fine-tuned DDPM on 10-shot FFHQ $\rightarrow$ Babies become more and more similar throughout training, as shown by the increasing cosine similarity.}
    \label{degrade}
\end{figure}

We provide the Nearest-LPIPS results of DDPMs trained from scratch on small-scale datasets in the top part of Table \ref{nearlpips}. For datasets containing 10 or 100 images, we have lower Nearest-LPIPS values. While for datasets containing 1000 images, we get measurably improved Nearest-LPIPS values. To avoid the influence of generated images symmetric to training samples, we flip all the training samples as supplements to the original datasets and recalculate the Nearest-LPIPS metric. The results are listed in the bottom part of Table \ref{nearlpips}. With the addition of flipped training samples, we find apparently lower Nearest-LPIPS values for datasets containing 10 or 100 images. However, we get almost the same Nearest-LPIPS results for DDPMs trained on larger datasets containing 1000 images, indicating that these models can generate diverse samples different from the original or symmetric training samples.

\begin{figure*}[t]
    \centering
    \includegraphics[width=1.0\linewidth]{ 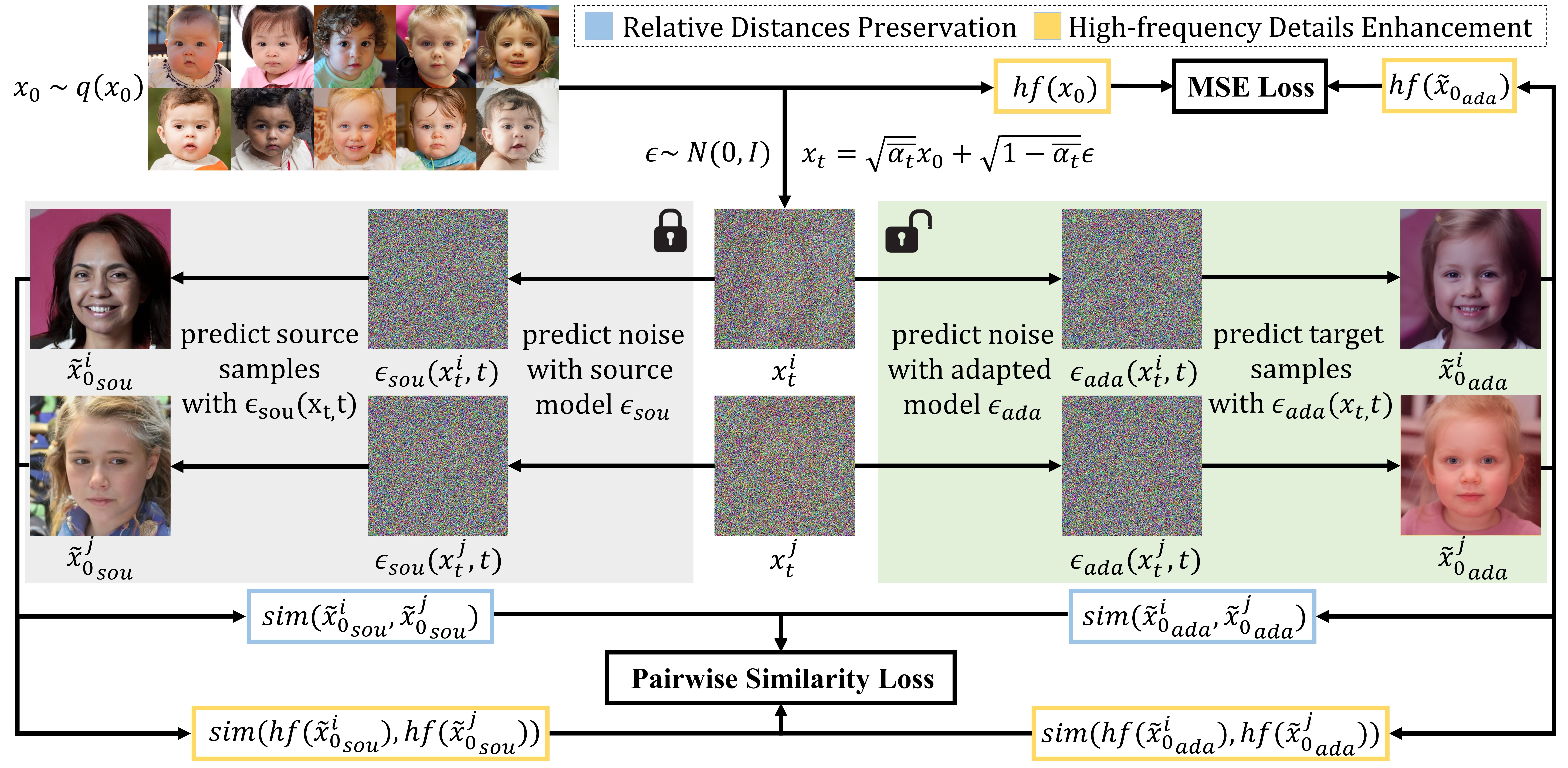}
    \caption{Overview of the DDPM-PA approach. A pairwise similarity loss is introduced to keep the relative pairwise distances of generated samples and their high-frequency details of adapted models similar to source models during adaptation. DDPM-PA guides adapted models to learn more high-frequency details from limited data using the MSE loss between high-frequency details extracted from training data and adapted samples. DDPM-PA generates high-quality and diverse samples sharing similar styles with few-shot training data.}
    \label{pairwise}
\end{figure*}

As proved by the results above, it becomes harder for DDPMs to learn the representations of datasets as training data become scarce. When trained on limited data from scratch, DDPMs fail to match target data distributions exactly or produce high-quality and diverse samples.

\section{Method}
To realize DDPM-based few-shot image generation, we first fine-tune DDPMs pre-trained on large-scale source datasets using limited target data directly. The fine-tuned models only need $3K$-$4K$ iterations to converge. As shown in the middle row of Fig. \ref{result1}, they can produce diverse results utilizing only 10 training samples. However, the generated samples lack ample high-frequency details and share similar features like hairstyles and facial expressions, leading to the degradation of generation quality and diversity.

Compared with pre-trained models, the degradation of fine-tuned models mainly comes from the excessively shortened relative distances between generated samples. As shown in Fig. \ref{degrade}, two samples synthesized from fixed noise inputs by the directly fine-tuned DDPM become increasingly similar (e.g., eyes and facial expressions) throughout training, losing various features and high-frequency details. Therefore, we propose to regularize the domain adaptation process by keeping the relative pairwise distances between adapted samples similar to source samples (Sec \ref{42}). Besides, we guide adapted models to learn high-frequency details from limited data and preserve high-frequency details learned from source domains (Sec \ref{44}). Our approach fixes source models as reference for adapted models. The weights of adapted models are initialized to the weights of source models and adapted to target domains. An overview of the proposed DDPM-PA approach is illustrated in Fig. \ref{pairwise} using 10-shot FFHQ $\rightarrow$ Babies as an example.

\begin{figure*}[t]
    \centering
    \includegraphics[width=1.0\linewidth]{ 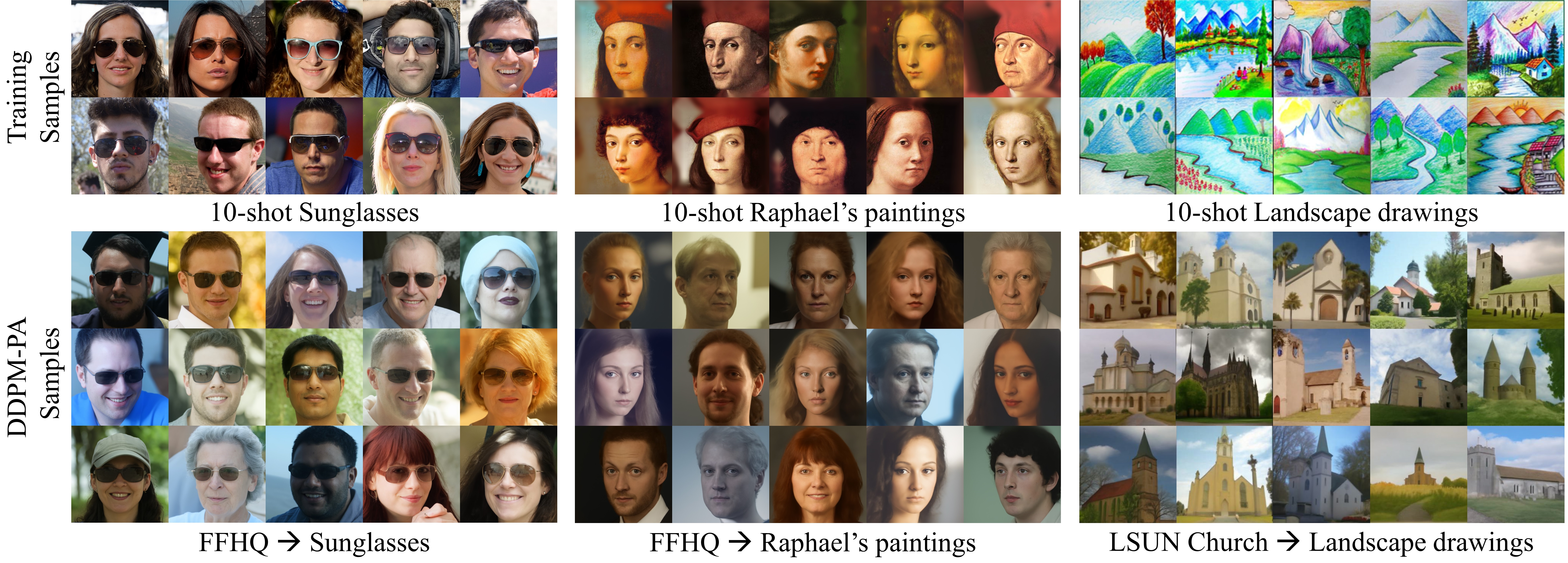}
    \caption{DDPM-PA samples on 10-shot FFHQ $\rightarrow$ Sunglasses, FFHQ $\rightarrow$ Raphael's paintings, and LSUN Church $\rightarrow$ Landscape drawings.}
    \label{result2}
\end{figure*}

\subsection{Relative Distances Preservation}
\label{42}
We design a pairwise similarity loss to preserve the relative distances between generated samples during domain adaptation. To construct N-way probability distributions for each image, we sample a batch of noised images $\left\lbrace x_t^{n} \right\rbrace_{n=0}^{N}$ by randomly adding Gaussian noises to training samples $x_0\sim q(x_0)$ following Equations \ref{eq1} and \ref{eq2}. Then the source and adapted models are applied to predict the fully denoised images $\left\lbrace \tilde{x}_0^{n} \right\rbrace_{n=0}^{N}$. We have the prediction of $\tilde{x}_0$ in terms of $x_t$ and $\epsilon_{\theta}(x_t,t)$ as follows:
\begin{align}
\label{eq11}
    \tilde{x}_0 = \frac{1}{\sqrt{\overline{\alpha}_t}}x_t-\frac{\sqrt{1-\overline{\alpha}_t}}{\sqrt{\overline{\alpha}_t}}\epsilon_{\theta}(x_t,t),
\end{align}
which can be derived from Equation \ref{eq4}. Cosine similarity is employed to measure the relative distances between the predicted samples $\tilde{x}_0$.  The probability distributions for $\tilde{x}_0^{i}\ (0\leq i \leq N)$ in the source and adapted models can be expressed as follows:
\begin{align}
    p_{i}^{sou} = sfm(\left\lbrace sim(\tilde{x}_{0_{sou}}^{i},\tilde{x}_{0_{sou}}^{j})\right\rbrace_{\forall i\neq j}), \label{16} \\ 
    p_{i}^{ada} = sfm(\left\lbrace sim(\tilde{x}_{0_{ada}}^{i},\tilde{x}_{0_{ada}}^{j})\right\rbrace_{\forall i\neq j}), \label{17}
\end{align}
where $sim$ and $sfm$ denote cosine similarity and softmax function, respectively. Then we have the pairwise similarity loss for generated images as follows:
\begin{align}
\label{18}
    \mathcal{L}_{img}(\epsilon_{sou},\epsilon_{ada}) = \mathbb{E}_{t,x_0,\epsilon} \sum_{i} D_{KL} (p_{i}^{ada}\,||\, p_{i}^{sou}),
\end{align}
where $D_{KL}$ represents KL-divergence.

\subsection{High-frequency Details Enhancement}
\label{44}

To begin with, we employ the typical Haar wavelet transformation \cite{0The} to disentangle images into multiple frequency components. Haar wavelet transformation contains four kernels including $LL^T$, $LH^T$, $HL^T$, $HH^T$, where $L$ and $H$ represent low and high pass filters, respectively:
\begin{align}
    L^T = \frac{1}{\sqrt{2}}[1,1], \quad H^T = \frac{1}{\sqrt{2}}[-1,1].
\end{align} 
Haar wavelet transformation decomposes inputs into four frequency components $LL$, $LH$, $HL$, and $HH$. $LL$ contains fundamental structures of images while other high-frequency components $LH$, $HL$, and $HH$ contain rich details of images. We define $hf$ as the sum of these high-frequency components:
\begin{align}
    hf = LH+HL+HH.
\end{align}

We implement high-frequency details enhancement from two perspectives. Firstly, we use the proposed pairwise similarity loss to preserve high-frequency details learned from source domains. Similarly, the probability distributions for the high-frequency components of $\tilde{x}_0^{i}\ (0\leq i \leq N)$ in the source and adapted models and the pairwise similarity loss for the high-frequency components in generated samples are as follows:
\begin{small}
\begin{align}
    & pf_{i}^{sou} = sfm(\left\lbrace sim(hf(\tilde{x}_{0_{sou}}^{i}),hf(\tilde{x}_{0_{sou}}^{j}))\right\rbrace_{\forall i\neq j}), \\
    & pf_{i}^{ada} = sfm(\left\lbrace sim(hf(\tilde{x}_{0_{ada}}^{i}),hf(\tilde{x}_{0_{ada}}^{j}))\right\rbrace_{\forall i\neq j}), \\
    & \mathcal{L}_{hf}(\epsilon_{sou},\epsilon_{ada}) = \mathbb{E}_{t,x_0,\epsilon} \sum_{i} D_{KL} (pf_{i}^{ada}\,||\, pf_{i}^{sou}),
\end{align}
\end{small}

Secondly, we guide adapted models to learn more high-frequency details from training data by minimizing the mean-squared error between the high-frequency components in adapted samples $\tilde{x}_0$ and training data $x_0$ as follows:
\begin{align}
    \mathcal{L}_{hfmse} = \mathbb{E}_{t,x_0,\epsilon} \left[||hf(\tilde{x}_0)-hf(x_0)||\right]^2.
\end{align}

\subsection{Overall Optimization Target}
The overall optimization target of the proposed DDPM-PA approach combines all the methods stated above: 
\begin{align}
\label{loss}
    \mathcal{L} = \mathcal{L}_{simple} + \lambda_1 \mathcal{L}_{vlb} + \lambda_2 \mathcal{L}_{img} +\lambda_3 \mathcal{L}_{hf} + \lambda_4 \mathcal{L}_{hfmse}.
\end{align}

We follow prior works \cite{nichol2021improved} to set $\lambda_1$ as 0.001 to avoid $\mathcal{L}_{vlb}$ from overwhelming $\mathcal{L}_{simple}$. $\mathcal{L}_{img}$ guides adapted models to preserve the relative distances between adapted samples. $\mathcal{L}_{hfmse}$ and $\mathcal{L}_{hf}$ further enhance the preservation of high-frequency details. We empirically find $\lambda_2,\lambda_3$ ranging between 0.1 and 1.0 and $\lambda_4$ ranging between 0.01 and 0.08 to be effective for few-shot adaptation setups.

\begin{figure*}[t]
    \centering
    \includegraphics[width=0.94\linewidth]{ 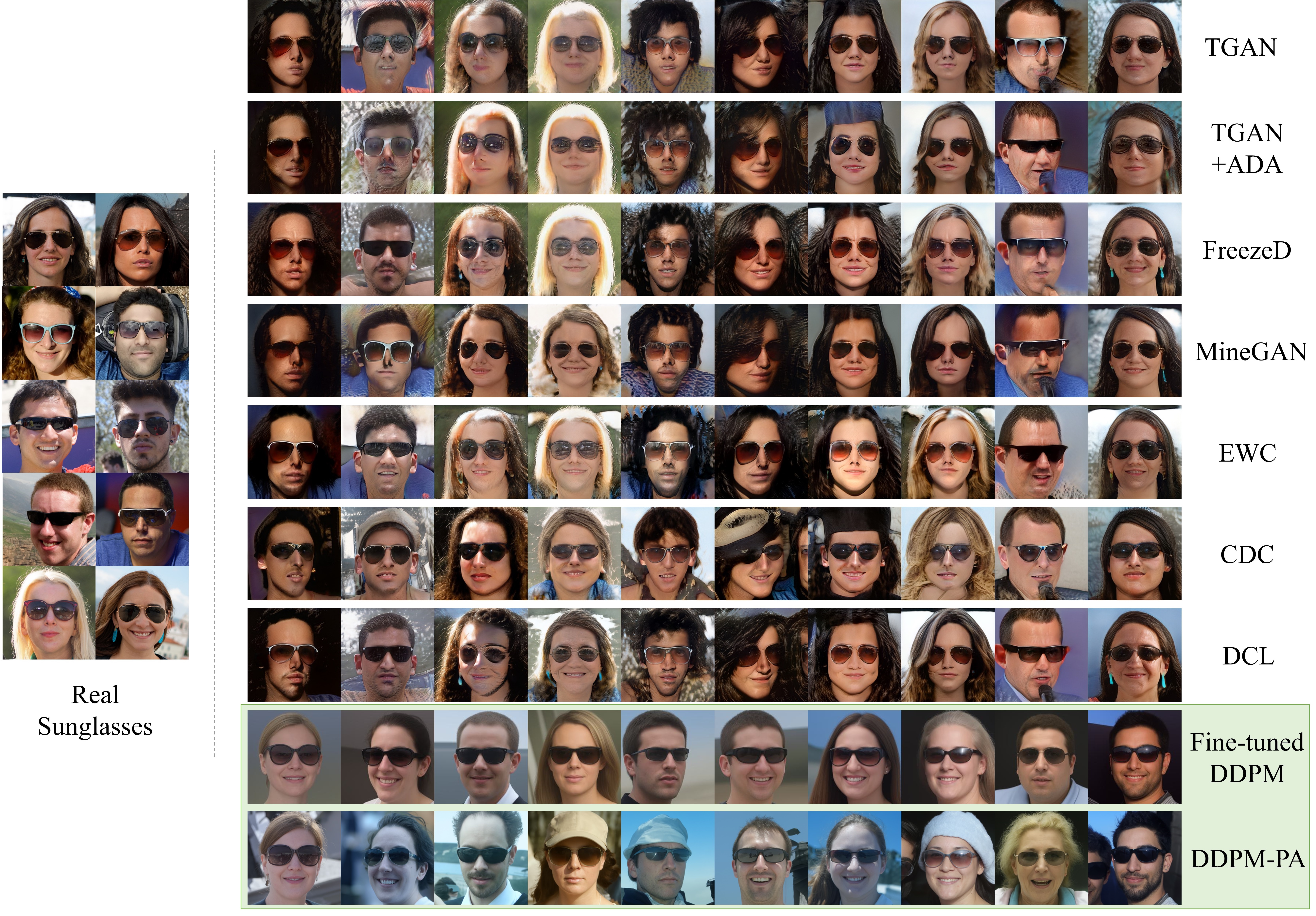}
    \caption{10-shot image generation samples on FFHQ $\rightarrow$ Sunglasses. All the samples of GAN-based approaches are synthesized from fixed noise inputs (rows 1-7). Samples of the directly fine-tuned DDPM and DDPM-PA are synthesized from fixed noise inputs as well (rows 8-9). Our approach generates high-quality results with fewer blurs and artifacts and achieves considerable generation diversity.}
    \label{sunglass}
\end{figure*}

\section{Experiments}
\label{422}

To demonstrate the effectiveness of DDPM-PA, we evaluate it with a series of few-shot image generation tasks using extremely few training samples (10 images). The performance of DDPM-PA is compared with directly fine-tuned DDPMs and modern GAN-based approaches on generation quality and diversity qualitatively and quantitatively.

\subsection{Quality and Diversity Evaluation}

\textbf{Basic Setups}  We choose FFHQ \cite{Karras_2020_CVPR} and LSUN Church \cite{yu2015lsun} as source datasets and train DDPMs from scratch on these two datasets for $300K$ and $250K$ iterations as source models. As for the target datasets, we employ 10-shot Sketches \cite{wang2008face}, Babies, Sunglasses \cite{Karras_2020_CVPR}, and face paintings by Amedeo Modigliani and Raphael Peale \cite{yaniv2019face} in correspondence to the source domain FFHQ. Besides, 10-shot Haunted houses and Landscape drawings are used as the target datasets in correspondence to LSUN Church. The model setups are consistent with the experiments on small-scale datasets in Sec. \ref{section3}. The DDPM-PA models are trained for $3K$-$5K$ iterations with a batch size of 24 on $\times 8$ NVIDIA RTX A6000 GPUs.

\begin{figure*}[t]
    \centering
    \includegraphics[width=1.0\linewidth]{ 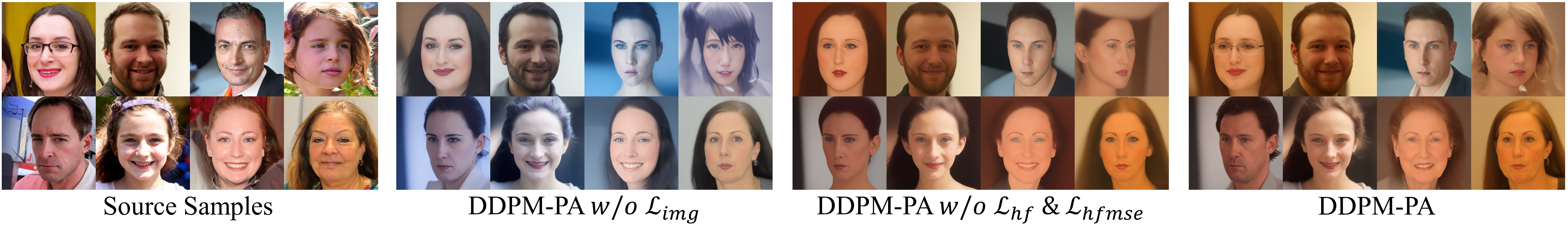}
    \caption{Visualized ablations analysis of the DDPM-PA approach using 10-shot FFHQ $\rightarrow$ Amedeo's paintings as an example.}
    \label{ablation}
\end{figure*}

\begin{table*}[tbp]
\centering
\setlength{\tabcolsep}{2mm}{
\begin{tabular}{l|c|c|c|c|c}
   Approaches & \makecell[c]{FFHQ $\rightarrow$ \\ Babies} & \makecell[c]{FFHQ $\rightarrow$ \\ Sunglasses} & \makecell[c]{FFHQ $\rightarrow$ \\ Raphael's paintings} & \makecell[c]{LSUN Church $\rightarrow$ \\ Haunted houses} & \makecell[c]{LSUN Church $\rightarrow$ \\ Landscape drawings} 
 \\
\hline
TGAN \cite{wang2018transferring} &  $0.510 \pm 0.026$ & $0.550 \pm 0.021$ & $0.533 \pm 0.023$ & $0.585 \pm 0.007$ & $0.601 \pm 0.030$ \\
TGAN+ADA \cite{ada} & $0.546 \pm 0.033$ & $0.571 \pm 0.034$ & $0.546 \pm 0.037$ &  $0.615 \pm 0.018$ & $0.643 \pm 0.060$ \\
FreezeD \cite{mo2020freeze} &  $0.535 \pm 0.021$ & $0.558 \pm 0.024$ & $0.537 \pm 0.026$ & $0.558 \pm 0.019$ & $0.597 \pm 0.032$ \\
MineGAN \cite{wang2020minegan} & $0.514 \pm 0.034$ & $0.570 \pm 0.020$ & $0.559 \pm 0.031$ & $0.586 \pm 0.041$ & $0.614 \pm 0.027$ \\
EWC \cite{ewc} & $0.560 \pm 0.019$ & $0.550 \pm 0.014$ & $0.541 \pm 0.023$ & $0.579 \pm 0.035$ & $0.596 \pm 0.052$ \\
CDC \cite{ojha2021few-shot-gan} & $0.583 \pm 0.014$ & $0.581 \pm 0.011$ & $0.564 \pm 0.010$ &  $0.620 \pm 0.029$ & $0.674 \pm 0.024$ \\
DCL \cite{zhao2022closer} & $0.579 \pm 0.018$ & $0.574 \pm 0.007$ & $0.558 \pm 0.033$ & $0.616 \pm 0.043$ & $0.626 \pm 0.021$ \\
\hline
Fine-tuned DDPMs & $0.513 \pm 0.026$ & $0.527 \pm 0.024$ & $0.466 \pm 0.018$ & $0.590 \pm 0.045$ &  $0.666 \pm 0.044$ \\
DDPM-PA (ours) & $\pmb{0.599 \pm 0.024}$ & $\pmb{0.604 \pm 0.014}$ & $\pmb{0.581 \pm 0.041}$ & $\pmb{0.628 \pm 0.029}$ & $\pmb{0.706 \pm 0.030 }$  \\
\end{tabular}}
\caption{Intra-LPIPS ($\uparrow$) results of DDPM-based approaches and GAN-based baselines on 10-shot image generation tasks adapted from the source domain FFHQ and LSUN Church. Standard deviations are computed across 10 clusters (the same number as training samples). DDPM-PA outperforms modern GAN-based approaches and achieves state-of-the-art performance in generation diversity.}
\label{intralpips}
\end{table*}

\textbf{Evaluation Metrics} We follow Ojha et al.'s work \cite{ojha2021few-shot-gan} to use Intra-LPIPS for generation diversity evaluation. To be more specific, we generate 1000 images and assign them to one of the training samples with the lowest LPIPS \cite{zhang2018unreasonable} distance. Intra-LPIPS is defined as the average pairwise LPIPS distances within members of the same cluster averaged over all the clusters. If a model exactly replicates training samples, its Intra-LPIPS will have a score of zero. Larger Intra-LPIPS values correspond to greater generation diversity. 

FID \cite{heusel2017gans} is widely used to evaluate the generation quality of generative models by computing the distribution distances between generated samples and datasets. However, FID would become unstable and unreliable when it comes to datasets containing a few samples (e.g., 10-shot datasets used in this paper). Therefore, we provide FID evaluation using relatively richer target datasets including Sunglasses and Babies, which contain 2500 and 2700 images.

We fix noise inputs for DDPM-based and GAN-based approaches respectively to synthesize samples for fair comparison of generation quality and diversity.

\textbf{Baselines}
Since few prior works realize few-shot image generation with DDPM-based approaches, we employ several GAN-based baselines sharing similar targets with us to adapt pre-trained models to target domains using only a few available samples for comparison: TGAN \cite{wang2018transferring}, TGAN+ADA \cite{ada}, FreezeD \cite{mo2020freeze}, MineGAN \cite{wang2020minegan}, EWC \cite{ewc}, CDC \cite{ojha2021few-shot-gan}, and DCL \cite{zhao2022closer}. All the methods are implemented based on the same StyleGAN2 \cite{Karras_2020_CVPR} codebase. DDPMs directly fine-tuned on limited data are included for comparison as well. The StyleGAN2 models and DDPMs trained on the large source datasets share similar generation quality and diversity (see more details in Appendix \ref{appendix_source}).

\textbf{Qualitative Evaluation}
We visualize the samples of DDPM-PA on 10-shot FFHQ $\rightarrow$ Babies, FFHQ $\rightarrow$ Sketches, and LSUN Church $\rightarrow$ Haunted houses in the bottom row of Fig. \ref{result1}. DDPM-PA produces more diverse samples containing richer high-frequency details than directly fine-tuned DDPMs. For example, DDPM-PA generates babies with various detailed hairstyles and facial features. Fig. \ref{result2} adds visualized samples under other adaptation setups. DDPM-PA adapts source models to target domains naturally and preserves various features different from training data. Samples of people wearing hats can be found when adapting FFHQ to Babies or Sunglasses, which is obviously different from the training samples. The adaptation from LSUN Church to Haunted houses and Landscape drawings retain various architectural structures. Fig. \ref{sunglass} shows samples of GAN-based and DDPM-based approaches on 10-shot FFHQ $\rightarrow$ Sunglasses. For intuitive comparison, we fix the noise inputs for GAN-based and DDPM-based approaches, respectively. GAN-based baselines generate samples containing unnatural blurs and artifacts. Besides, we find many incomplete sunglasses in the generated samples, leading to vague visual effects. In contrast, the directly fine-tuned DDPM produces smoother results but lacks details. DDPM-PA achieves more diverse and realistic samples containing richer details than existing approaches. Additional visualized comparison results can be found in Appendix \ref{appendix_results}.%

\begin{table}[t]
\centering
\setlength\tabcolsep{3pt}
\small
\begin{tabular}{l|c|c|c|c|c|c}
Method & TGAN & ADA & EWC & CDC & DCL & DDPM-PA\\
\hline
Babies & $104.79$ & $102.58$ & $87.41$ & $74.39$ & $52.56$ & $ \pmb{48.92}$ \\
Sunglasses & $55.61$ & $53.64$ & $59.73$ & $42.13$ & $38.01$ & $ \pmb{34.75}$\\
\end{tabular}
\caption{FID ($\downarrow$) results of DDPM-PA compared with GAN-based baselines on 10-shot FFHQ $\rightarrow$ Babies and Sunglasses.}
\label{fid}
\end{table}

\textbf{Quantitative Evaluation}
We provide the Intra-LPIPS results of DDPM-PA under a series of 10-shot adaptation setups in Table \ref{intralpips}. DDPM-PA realizes a superior improvement of Intra-LPIPS compared with directly fine-tuned DDPMs. Besides, DDPM-PA outperforms state-of-the-art GAN-based approaches, indicating its strong capability of maintaining generation diversity. Intra-LPIPS results under other adaptation setups are added in Appendix \ref{appendix_fid}. 

As shown by the FID results in Table \ref{fid}, DDPM-PA also performs better on learning target distributions from limited data than prior GAN-based approaches.

\subsection{Ablation Analysis}
\label{43}
We ablate our approach using 10-shot FFHQ $\rightarrow$ Amedeo's paintings as an example. With high-frequency details enhancement only, the adapted model cannot match target distributions and produces samples containing unnatural artifacts. With relative distances preservation only, the adapted model produces coarse samples lacking details like hairstyles. DDPM-PA combines both ideas and obtains realistic and diverse results. Ablations of the weight coefficients in Equation \ref{loss} are added in Appendix \ref{appendix_ablation}.

\section{Conclusion}
This work first introduces diffusion models to few-shot image generation. However, when trained on limited data, DDPMs can only replicate training samples or produce coarse results. Therefore, we propose the DDPM-PA approach to adapt source models to target domains while maintaining diversity and preserving details. DDPM-PA keeps the relative pairwise distances between adapted samples and realizes high-frequency details enhancement during domain adaptation. We demonstrate the effectiveness of DDPM-PA on a series of few-shot image generation tasks. It generates compelling samples with rich details and few blurs, outperforming current state-of-the-art GAN-based approaches on generation quality and diversity. Our work is an essential step toward more data-efficient diffusion models. The limitations are discussed in Appendix \ref{appendix_limitations}.

\bibliographystyle{abbrv}
\bibliography{refs}

\clearpage

\appendix

\section{Supplementary DPPM-PA Evaluation}
\label{appendix_fid}

In Table \ref{intralpips2}, we provide the additional results of Intra-LPIPS \cite{ojha2021few-shot-gan} on 10-shot FFHQ $\rightarrow$ Sketches and FFHQ $\rightarrow$ Amedeo's paintings as supplements to Table \ref{intralpips}. The proposed DDPM-PA approach outperforms modern GAN-based approaches and directly fine-tuned DDPMs on generation diversity under these adaptation setups as well.

FID results of GAN-based baselines in Table \ref{fid} are reported in CDC \cite{ojha2021few-shot-gan} and DCL \cite{zhao2022closer}.

\section{Supplementary Ablation Analysis}
\label{appendix_ablation}
 In this section, we provide the ablation analysis of the weight coefficients of $\mathcal{L}_{img}$, $\mathcal{L}_{hf}$, and $\mathcal{L}_{hfmse}$ using 10-shot FFHQ $\rightarrow$ Babies as an example. Intra-LPIPS and FID are employed for quantitative evaluation. 
 
\begin{figure}[ht]
    \centering
    \includegraphics[width=1.0\linewidth]{ 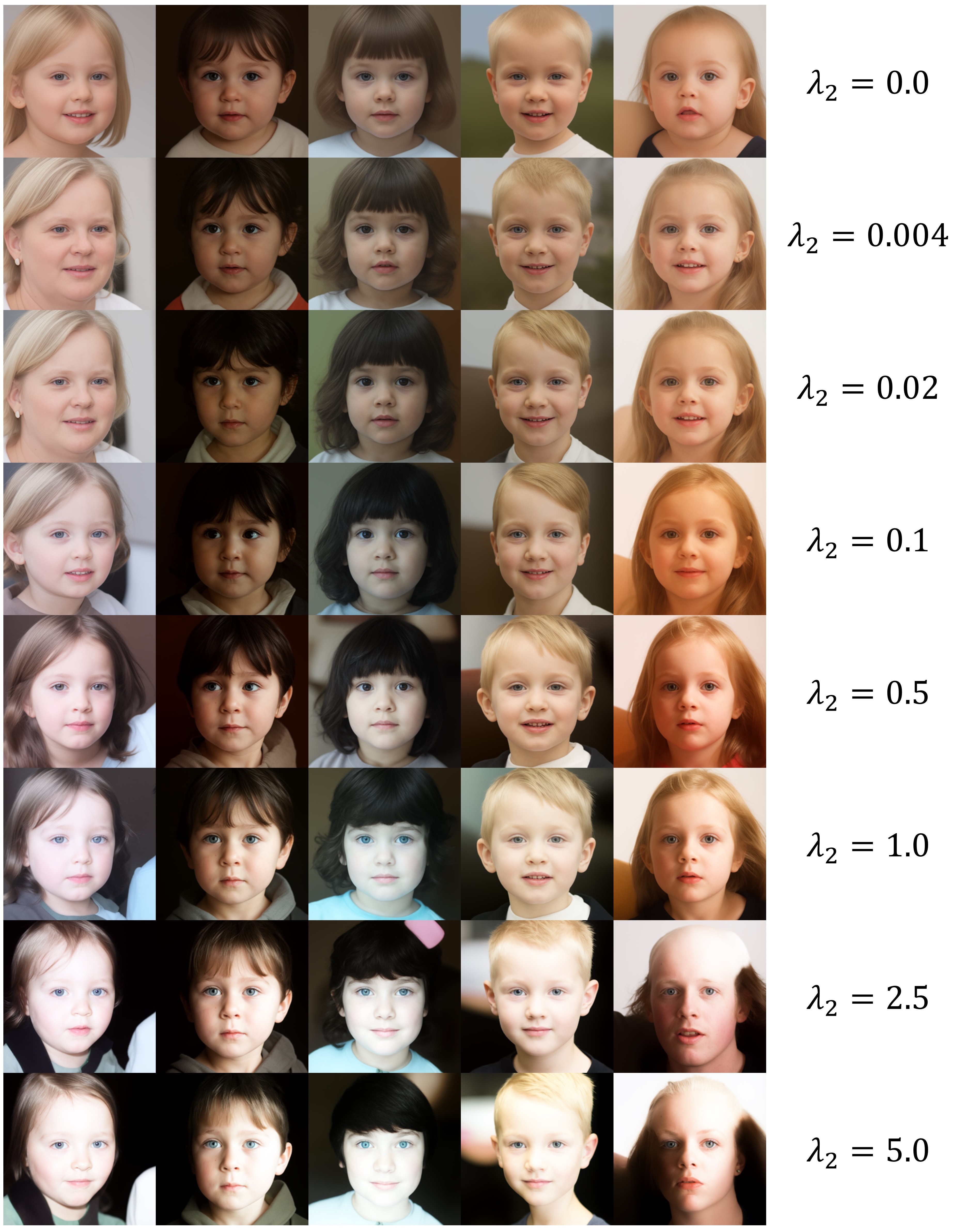}
    \caption{Visualized ablations of $\lambda_2$, the weight coefficient of $\mathcal{L}_{img}$ on 10-shot FFHQ $\rightarrow$ Babies. Samples of different models are synthesized from fixed noise inputs.}
    \label{lambda2_img}
\end{figure}

We first ablate $\lambda_2$, the weight coefficient of $\mathcal{L}_{img}$. We adapt the source model to 10-shot Babies without $\mathcal{L}_{hf}$ and $\mathcal{L}_{hfmse}$. The quantitative results are listed in Table \ref{lambda2}. Corresponding generated samples are shown in Fig. \ref{lambda2_img}. When $\lambda_2$ is set as 0.0, the directly fine-tuned model produces coarse results lacking high-frequency details and diversity. With an appropriate choice of $\lambda_2$, the adapted model achieves greater generation diversity and better learning of target distributions under the guidance of $\mathcal{L}_{img}$. Too large values of $\lambda_2$ make $\mathcal{L}_{img}$ overwhelm $\mathcal{L}_{simple}$ and prevent the adapted model from learning target distributions, leading to degraded generation quality and diversity. The adapted model with $\lambda_2$ value of 2.5 gets unnatural generated samples even if it achieves the best FID result. We recommend $\lambda_2$ ranging from $0.1$ to $1.0$ for the adaptation setups used in our paper based on a comprehensive consideration of the qualitative and quantitative evaluation.

\begin{figure}[ht]
    \centering
    \includegraphics[width=1.0\linewidth]{ 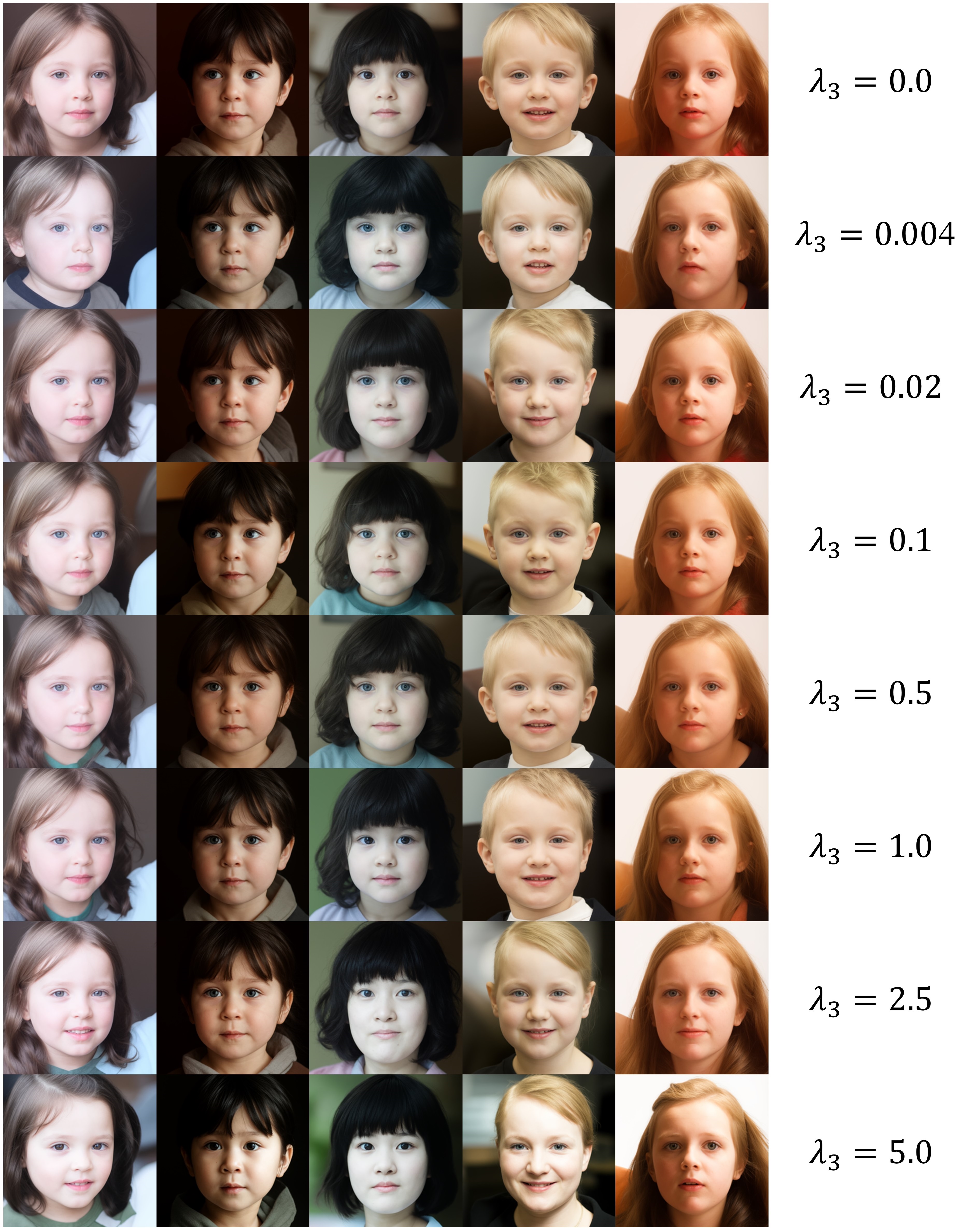}
    \caption{Visualized ablations of $\lambda_3$, the weight coefficient of $\mathcal{L}_{hf}$ on 10-shot FFHQ $\rightarrow$ Babies. Samples of different models are synthesized from fixed noise inputs.}
    \label{lambda3_img}
\end{figure}

Next, we ablate $\lambda_3$, the weight coefficient of $\mathcal{L}_{hf}$ with $\lambda_2$ set as 0.5. The quantitative results are listed in Table \ref{lambda3}. Corresponding generated samples are shown in Fig. \ref{lambda3_img}. $\mathcal{L}_{hf}$ guides adapted models to keep diverse high-frequency details learned from source domains for more realistic results. $\mathcal{L}_{hf}$ helps the adapted model enhance details like clothes and hairstyles and achieves better FID and Intra-LPIPS, indicating improved quality and diversity. Too large values of $\lambda_3$ make the adapted model pay too much attention to high-frequency components and fail to produce realistic results following the target distributions. We recommend $\lambda_3$ ranging from 0.1 to 1.0 for the adaptation setups used in our paper.

\begin{table}[t]
\centering
\small
\begin{tabular}{l|c|c}
 Approaches & \makecell[c]{FFHQ $\rightarrow$ \\ Sketches} & \makecell[c]{FFHQ $\rightarrow$ \\ Amedeo's paintings} 
 \\
 \hline
 TGAN \cite{wang2018transferring} &  $0.394 \pm 0.023$ & $0.548 \pm 0.026$  \\
 TGAN+ADA \cite{ada} & $0.427 \pm 0.022$ & $0.560 \pm 0.019$  \\
 FreezeD \cite{mo2020freeze} & $0.406 \pm 0.017$ & $0.597 \pm 0.032$  \\
 MineGAN \cite{wang2020minegan} & $0.407 \pm 0.020$ & $0.614 \pm 0.027$  \\
 EWC \cite{ewc} & $0.430 \pm 0.018$ & $0.594 \pm 0.028$  \\
 CDC \cite{ojha2021few-shot-gan} &  $0.454 \pm 0.017$ & $0.620 \pm 0.029$  \\
 DCL \cite{zhao2022closer} & $0.461 \pm 0.021$ & $0.616 \pm 0.043$  \\
 \hline
 Fine-tuned DDPMs  & $0.473 \pm 0.022$ & $0.484 \pm 0.021$ \\
 DDPM-PA (ours) &  $\pmb{0.495 \pm 0.024}$ & $\pmb{0.626 \pm 0.022}$ \\
\end{tabular}
\caption{Intra-LPIPS ($\uparrow$) results of DDPM-based approaches and GAN-based baselines on 10-shot FFHQ $\rightarrow$ Sketches and FFHQ $\rightarrow$ Amedeo's paintings. Standard deviations are computed across 10 clusters (the same number as training samples).}
\label{intralpips2}
\end{table}

\begin{table}[tbp]
    \centering
    \begin{tabular}{c|c|c}
        $\lambda_2$ & Intra-LPIPS ($\uparrow$) & FID ($\downarrow$) \\
        \hline
        $0.0$ &  $0.520 \pm 0.026$ & $114.95$ \\
        $0.004$ & $0.531 \pm 0.031$ & $92.87$ \\
        $0.02$ & $0.544 \pm 0.026$ & $85.11$\\
        $0.1$ & $0.558 \pm 0.033$ & $75.17$\\
        $0.5$ & $\pmb{0.572 \pm 0.027}$ & $71.77$\\
        $1.0$ & $0.560 \pm 0.034$ & $74.68$\\
        $2.5$ & $0.543 \pm 0.038$ & $\pmb{64.08}$\\
        $5.0$ & $0.537 \pm 0.028$ & $69.18$\\
    \end{tabular} 
    \caption{Intra-LPIPS ($\uparrow$) and FID ($\downarrow$) results of adapted models trained on 10-shot FFHQ $\rightarrow$ Babies with different $\lambda_2$, the weight coefficient of $\mathcal{L}_{img}$.}
    \label{lambda2}
\end{table}

\begin{table}[tbp]
    \centering
    \begin{tabular}{c|c|c}
        $\lambda_3$ & Intra-LPIPS ($\uparrow$) & FID ($\downarrow$) \\
        \hline
        $0.0$ & $0.572 \pm 0.027$ & $71.77$ \\
        $0.004$ & $0.576 \pm 0.034$ & $\pmb{66.48}$ \\
        $0.02$ & $0.581 \pm 0.045$ & $72.67$ \\
        $0.1$ & $0.589 \pm 0.047$ & $70.75$ \\
        $0.5$ & $\pmb{0.592 \pm 0.031}$ & $70.40$ \\
        $1.0$ & $0.583 \pm 0.032$ & $68.06$ \\
        $2.5$ & $0.577 \pm 0.032 $ & $71.69$ \\
        $5.0$ & $0.591 \pm 0.031 $ & $71.20$ \\
    \end{tabular} 
    \caption{Intra-LPIPS ($\uparrow$) and FID ($\downarrow$) results of adapted models trained on 10-shot FFHQ $\rightarrow$ Babies with different $\lambda_3$, the weight coefficient of $\mathcal{L}_{hf}$.}
    \label{lambda3}
\end{table}

\begin{figure}[ht]
    \centering
    \includegraphics[width=1.0\linewidth]{ 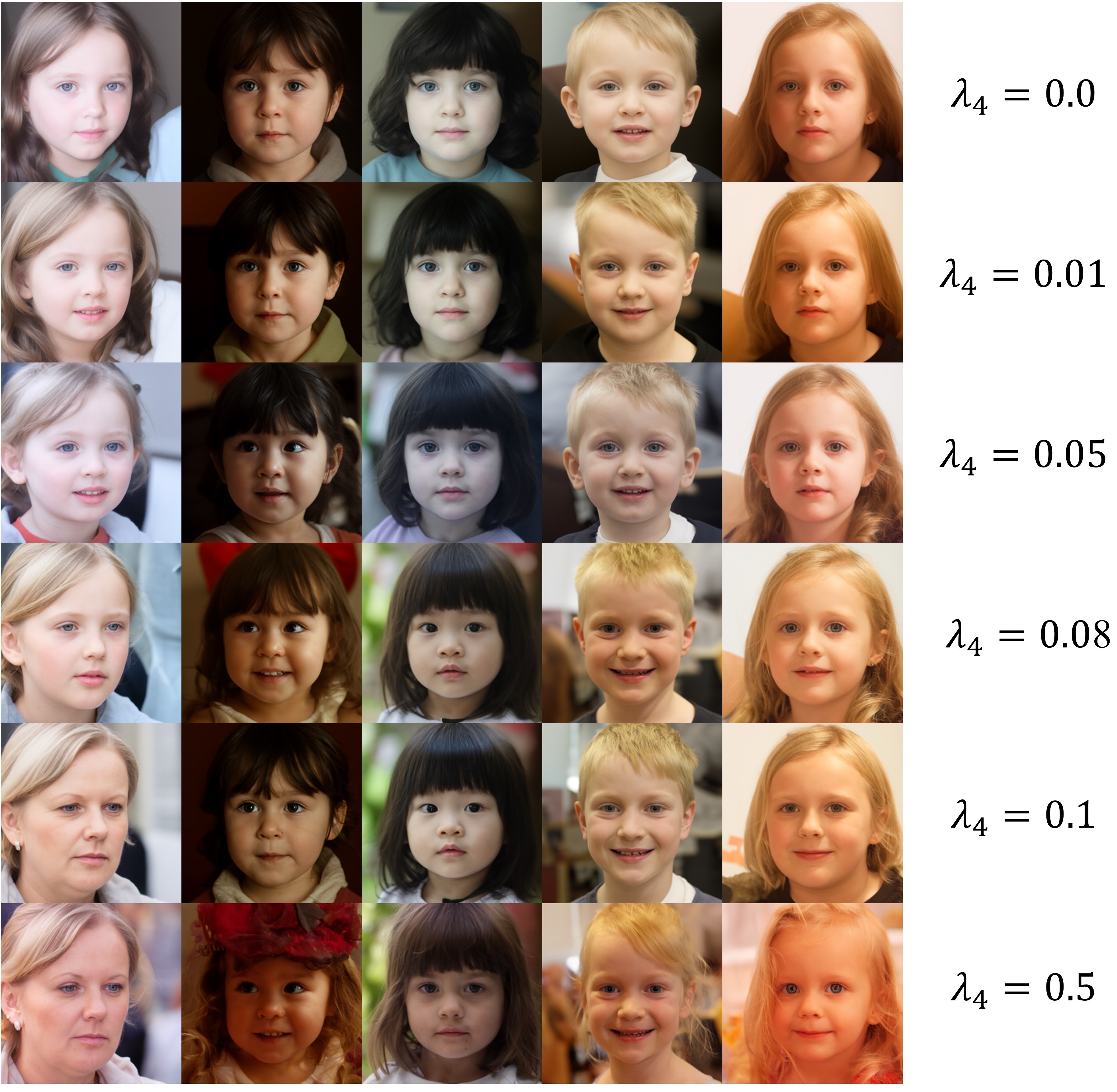}
    \caption{Visualized ablations of $\lambda_4$, the weight coefficient of $\mathcal{L}_{hfmse}$ on 10-shot FFHQ $\rightarrow$ Babies. Samples of different models are synthesized from fixed noise inputs.}
    \label{lambda4_img}
\end{figure}

\begin{table}[tbp]
    \centering
    \begin{tabular}{c|c|c}
        $\lambda_4$ & Intra-LPIPS ($\uparrow$) & FID ($\downarrow$) \\
        \hline
        $0.0$ & $0.592 \pm 0.031$ & $70.40$ \\
        $0.01$ & $0.594 \pm 0.038$ & $66.31$ \\
        $0.05$ & $0.599 \pm 0.024$ & $\pmb{48.92}$\\
        $0.08$ & $0.607 \pm 0.025$ & $ 55.88 $\\
        $0.1$ & $0.603 \pm 0.031$ &  $59.28$\\
        $0.5$ & $\pmb{0.612 \pm 0.023}$ & $70.26$\\
    \end{tabular} 
    \caption{Intra-LPIPS ($\uparrow$) and FID ($\downarrow$) results of adapted models trained on 10-shot FFHQ $\rightarrow$ Babies with different $\lambda_4$, the weight coefficient of $\mathcal{L}_{hfmse}$.}
    \label{lambda4}
\end{table}

Finally, we ablate $\lambda_4$, the weight coefficient of $\mathcal{L}_{hfmse}$, with $\lambda_2$ and $\lambda_3$ set as 0.5. The quantitative results are listed in Table \ref{lambda4}. Corresponding generated samples are shown in Fig. \ref{lambda4_img}. $\mathcal{L}_{hfmse}$ guides the adapted model to learn more high-frequency details from limited training data. Appropriate choice of $\lambda_4$ helps the adapted model generate diverse results containing rich details. Besides, the full DDPM-PA approach achieves state-of-the-art results of FID and Intra-LPIPS on 10-shot FFHQ $\rightarrow$ Babies (see Table \ref{intralpips} and \ref{fid}). Similar to $\lambda_2$ and $\lambda_3$, too large values of $\lambda_4$ lead to unreasonable results deviating from the target distributions. We recommend $\lambda_4$ ranging from 0.01 to 0.08 for the adaptation setups in this paper.

\begin{figure*}[ht]
    \centering
    \includegraphics[width=1.0\linewidth]{ 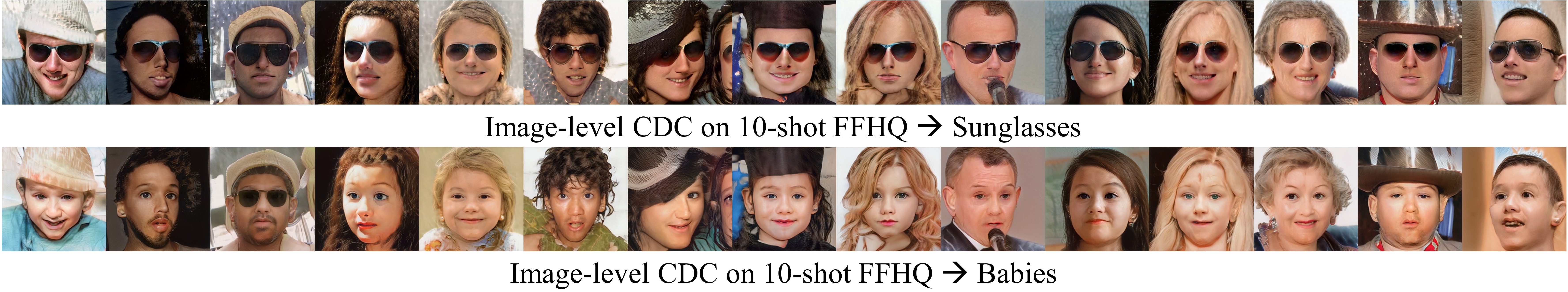}
    \caption{Image samples synthesized by CDC \cite{ojha2021few-shot-gan} using image-level information on 10-shot FFHQ $\rightarrow$ Sunglasses and FFHQ $\rightarrow$ Babies. }
    \label{imagecdc}
\end{figure*}

\section{Limitations}
\label{appendix_limitations}
 Despite the compelling results of our approach, it still has some limitations. All the datasets used in this paper have the resolution of $256\times 256$. The experiments of DDPM-PA are conducted on $\times 8$ NVIDIA RTX A6000 GPUs (48 GB memory of each). However, the batch size on each GPU is still limited to 3. Therefore, it is challenging to expand our approach to larger image resolution. We will work on more lightweight DDPM-based few-shot image generation approaches. Despite that, the datasets used in this paper have larger resolution than many DDPM-based works \cite{giannone2022few,nichol2021improved,austin2021structured,chen2022analog,kingma2021variational,zhang2022gddim} which use datasets with resolution $32\times 32$ and $64\times 64$.

Although we have designed methods for high-frequency details enhancement and achieved realistic results, there still exists room for further improvement, especially when target domains contain apparently more high-frequency components than source domains (e.g., LSUN Church $\rightarrow$ Haunted houses). Besides, DDPM-PA still cannot fully reproduce the styles of some abstract target domains while maintaining generation diversity. Nevertheless, our work first introduces diffusion models to few-shot image generation tasks and has improved generation quality and diversity compared with existing GAN-based approaches. We hope it will be a solid basis for better methods in the future.

\section{Inspiration of Loss Design}
\subsection{$\mathcal{L}_{img} \; \& \; \mathcal{L}_{hf}$}
The proposed pairwise similarity loss designed for DDPMs is mainly inspired by the methods in contrastive learning \cite{oord2018representation, he2020momentum, chen2020simple}, which build probability distributions based on similarities. Similar methods are applied to GAN-based approaches, including CDC \cite{ojha2021few-shot-gan} and DCL \cite{zhao2022closer} as well. GAN-based approaches depend on perceptual features in the generator and discriminator to compute similarity and probability distributions. As for the proposed DDPM-PA approach, the predicted input images $\tilde{x}_0$ calculated in terms of $x_t$ and $\epsilon_{\theta}(x_t,t)$ (Equation \ref{eq11}) are applied in replacement of perceptual features used for GANs. DDPM-PA directly uses image-level information to preserve the relative pairwise distances between adapted samples during domain adaptation. 

We tried to use features in diffusion processes (Design A) and images of several diffusion steps (Design B) for pairwise similarity loss calculation. As shown in Table \ref{compar} (FID evaluation on FFHQ $\rightarrow$ Sunglasses, Intra-LPIPS evaluation on 10-shot FFHQ $\rightarrow$ Sunglasses), the proposed loss design is simple, effective, inexpensive, and achieves the best quality and diversity. Here we do not include high-frequency details enhancement for fair comparison.

\begin{table}[t]
\centering
\small
\begin{tabular}{l|c|c|c}
    Method & FID ($\downarrow$) & Intra-LPIPS ($\uparrow$) & Time $/\ 1K$ iters ($\downarrow$) \\
    \hline
    Ours & $\pmb{37.92}$ & $\pmb{0.59 \pm 0.02}$ & $\pmb{34 min}$ \\ 
    Design A & $40.30$ & $0.55 \pm 0.03$ & $52 min$\\ 
    Design B & $58.28$ & $0.57 \pm 0.06$ & $ 38 min$ \\
\end{tabular} 
\caption{Quantitative evaluation comparison between different loss designs.}
\label{compar}
\end{table}

As illustrated in Sec. \ref{422}, DDPM-PA synthesizes more realistic images with fewer blurs and artifacts and achieves better generation diversity than current state-of-the-art GAN-based approaches \cite{ojha2021few-shot-gan, zhao2022closer}. We also try to use image-level information to replace the perceptual features for the GAN-based approach CDC \cite{ojha2021few-shot-gan}. However, we fail to avoid generating artifacts or achieve higher generation quality, as shown in Fig. \ref{imagecdc}. The proposed pairwise similarity loss matches better with diffusion models than GANs. 

\subsection{$\mathcal{L}_{hfmse}$}

DDPMs learn target distributions mainly through mean values of predicted noises using the reweighted loss function (Equation \ref{loss_simple}). As a result, it is hard for DDPMs to learn high-frequency distributions from limited data, as shown in the smooth samples produced by models trained on limited data from scratch in Fig. \ref{scratch2}. Therefore, we propose $\mathcal{L}_{hfmse}$ to strengthen the learning of high-frequency details from limited data during domain adaptation.

\begin{figure*}[ht]
    \centering
    \includegraphics[width=1.0\linewidth]{ }
    \caption{Examples of small-scale datasets sampled from Babies, Sunglasses, and LSUN Church. For datasets containing 100 or 1000 images, we randomly pick 15 examples.}
    \label{scratch_data}
\end{figure*}

\section{More Details of Implementation}
\label{appendix_implementation}
\subsection{More Details of DDPMs} 
We follow the model setups of DDPMs used in prior works \cite{nichol2021improved} for LSUN $256^2$ \cite{yu2015lsun} datasets. All the DDPM-based models used in this paper are implemented based on the same codebase \cite{nichol2021improved,dhariwal2021diffusion} and share the same model structure for fair comparison under different adaptation setups and optimization targets. All the source and target datasets are modified to the resolution of $256 \times 256$. We use a max diffusion step $T$ of 1000 and a dropout rate of 0.1. The models are trained to learn the variance with $\mathcal{L}_{vlb}$. The Adam optimizer \cite{kingma2014adam} is employed to update the trainable parameters. We set the learning rate as 0.001 and apply the linear noise addition schedule to the noising process. Besides, we use half-precision (FP16) binary floating-point format to save memory and make it possible to use a larger batch size in our experiments (batch size 6 for directly fine-tuned DDPMs and batch size 3 for DDPM-PA per NVIDIA RTX A6000 GPU). All the results produced by DDPM-based models in this paper follow the sampling process proposed in Ho et al.'s work \cite{NEURIPS2020_4c5bcfec} (about 21 hours needed to generate 1000 samples on a single NVIDIA RTX A6000 GPU) without any fast sampling methods \cite{song2020denoising, zhang2022gddim, lu2022dpm, lu2022dpm2, zhang2022fast, karras2022elucidating}. The weight coefficient $\lambda_2$, $\lambda_3$, and $\lambda_4$ are set as 0.5, 0.5, 0.05 for the results of DDPM-PA listed in Table \ref{intralpips}, \ref{fid}, and \ref{intralpips2}.

\begin{figure*}[tbp]
    \centering
    \includegraphics[width=1.0\linewidth]{ }
    \caption{Image samples produced by DDPMs trained from scratch on small-scale datasets, including Babies, Sunglasses, and LSUN Church containing 10, 100, and 1000 images. }
    \label{scratch2}
\end{figure*}

\subsection{More Details of GAN-based Baselines}
We employ several GAN-based few-shot image generation approaches as baselines for comparison with the proposed DDPM-PA approach. Here we provide more details of these baselines. We implement all these approaches based on the same codebase of StyleGAN2 \cite{Karras_2020_CVPR}. The source models are fine-tuned directly on the target datasets to realize TGAN \cite{wang2018transferring}. TGAN+ADA applies ADA \cite{ada} augmentation method to the TGAN baseline. For FreezeD \cite{mo2020freeze}, the first 4 high-resolution layers of the discriminator are frozen following the ablations analysis provided in their work. The results of MineGAN \cite{wang2020minegan} and CDC \cite{ojha2021few-shot-gan} are produced through their official implementation. As for EWC \cite{ewc} and DCL \cite{zhao2022closer}, we implement these approaches following formulas and parameters in their papers since there is no official implementation. These GAN-based approaches are designed for generators \cite{wang2020minegan,ewc,ojha2021few-shot-gan,zhao2022closer} and discriminators \cite{ada,mo2020freeze,zhao2022closer} specially and cannot be expanded to DDPMs directly.

  \begin{figure}[t]
    \centering
    \includegraphics[width=1.0\linewidth]{ 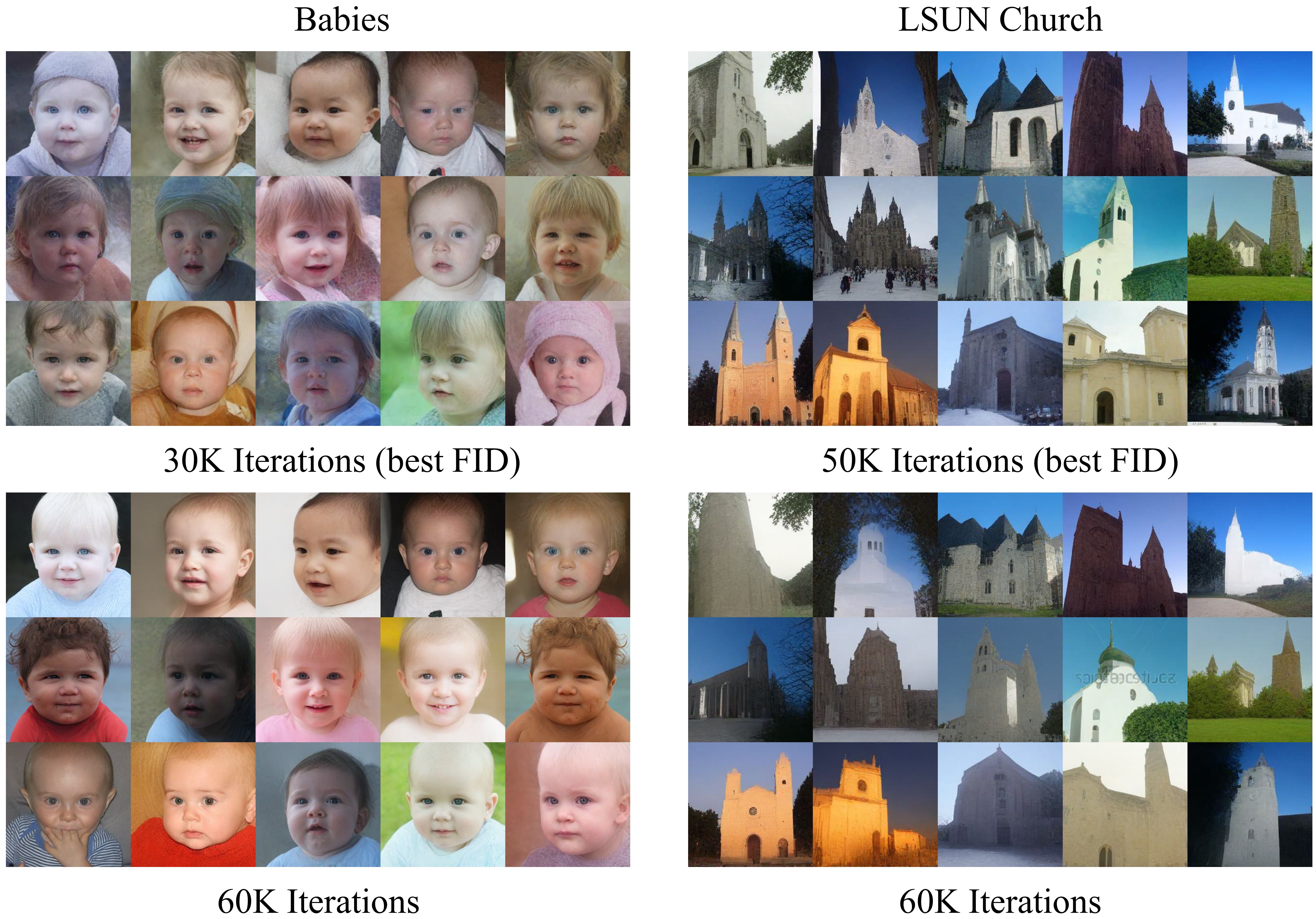}
    \caption{Visualized comparison between DDPMs that achieve the best FID results and DDPMs trained for $60K$ iterations on 1000-shot Babies and LSUN Church. Image samples produced by different models are synthesized from fixed noise inputs.}
    \label{scratch_compare}
\end{figure}

\begin{table}[tbp]
    \centering
    \begin{tabular}{c|c|c|c}
        Iterations (K) & Babies & Sunglasses & \makecell[c]{LSUN Church}  \\
        \hline
        $0$ & $444.35$ & $419.75$ & $424.53$ \\
        $10$ & $444.91$ & $419.38$ & $413.68$ \\
        $20$ & $222.81$ & $348.48$ & $385.25$ \\
        $30$ & $\pmb{90.16}$ & $168.62$ & $388.81$ \\
        $40$ & $124.97$ & $82.48$ & $57.68$ \\
        $50$ & $132.33$ & $68.26$ & $\pmb{45.43}$ \\
        $60$ & $132.32$ & $\pmb{66.09}$ & $69.18$ \\
    \end{tabular} 
    \caption{FID ($\downarrow$) results of DDPMs trained for different iterations from scratch on 1000-shot Babies, Sunglasses, and LSUN Church.}
    \label{fid_table}
\end{table}

The adapted GANs are trained for $1K$-$3K$ iterations. When evaluating generation diversity using Intra-LPIPS \cite{ojha2021few-shot-gan} for the 10-shot adaptation tasks listed in Tables \ref{intralpips} and \ref{intralpips2}, we apply fixed noise inputs to different GAN-based approaches for fair comparison. 

\begin{figure*}[tbp]
    \centering
    \includegraphics[width=1.0\linewidth]{ 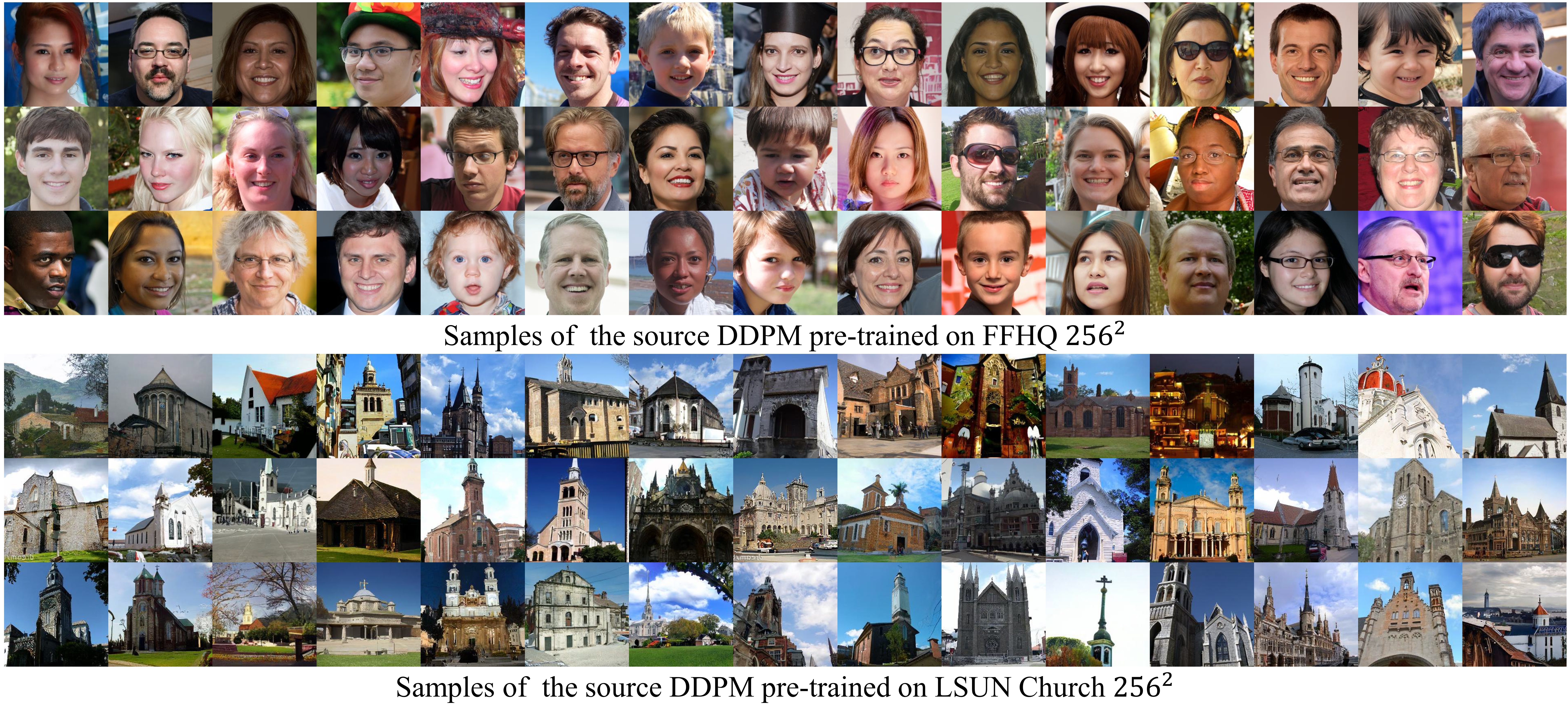}
    \caption{Image samples produced by DDPMs trained on FFHQ $256^2$ \cite{Karras_2020_CVPR} ($300K$ iterations) and LSUN Church $256^2$ \cite{yu2015lsun} ($250K$ iterations).}
    \label{result_scratch}
\end{figure*}

Apart from the GAN-based baselines illustrated above, RSSA \cite{xiao2022few} and AdAM \cite{adaptative} provide different research perspectives for few-shot image generation. RSSA \cite{xiao2022few} preserves the image structure learned from source domains with a relaxed spatial structural alignment method, which is inappropriate for abstract target domains like artists' paintings. AdAM \cite{adaptative} focuses on unrelated source/target domains with an adaptation-aware kernel modulation approach.

\section{DDPMs Trained from Scratch}
\label{appendix_scratch}
In Section \ref{section3}, we evaluate the performance of DDPMs trained from scratch on small-scale datasets containing 10, 100, and 1000 images. In our experiments, the smaller datasets are included in the larger datasets. For example, 1000-shot Sunglasses includes all the images in 100-shot and 10-shot Sunglasses. Similarly, all the images in 10-shot Sunglasses are included in 100-shot Sunglasses as well. We train DDPMs for $40K$ iterations (about 20 hours on $\times 8$ NVIDIA RTX A6000 GPUs) on datasets containing 10 or 100 images. While for datasets containing 1000 images, DDPMs are trained for $60K$ iterations (about 30 hours on $\times 8$ NVIDIA RTX A6000 GPUs).

 We provide several typical examples randomly picked from the small-scale datasets in Fig. \ref{scratch_data}. Compared with the generated images shown in Fig. \ref{scratch2}, it can be seen that DDPMs trained from scratch need enough training samples (e.g., 1000 images) to synthesize diverse results and avoid replicating the training samples. Detailed FID \cite{heusel2017gans} results of DDPMs trained from scratch on Babies, Sunglasses, and LSUN Church containing 1000 images for $60K$ iterations are shown in Table \ref{fid_table}. DDPMs trained from scratch on 1000-shot Babies, Sunglasses, and LSUN Church achieve the best FID results at $30K$, $60K$, and $50K$ iterations, respectively.
 
 \begin{table}[tbp]
    \centering
    \begin{tabular}{l|c|c}
        Approaches & FFHQ & LSUN Church  \\
        \hline
        StyleGAN2 & $0.6619 \pm 0.0581$   &  $ 0.7144 \pm 0.0537$ \\
        DDPM &   $\pmb{0.6631 \pm 0.0592}$   & $\pmb{0.7153 \pm 0.0513}$
    \end{tabular} 
    \caption{Average pairwise LPIPS ($\uparrow$) results of 1000 samples produced by StyleGAN2 and DDPMs trained on FFHQ $256^2$ and LSUN Church $256^2$.}
    \label{source_lpips}
\end{table}

\begin{table}[tbp]
    \centering
    \begin{tabular}{l|c|c}
        Approaches & FFHQ & LSUN Church  \\
        \hline
        StyleGAN2 &  $7.71$  & $8.09$   \\
        DDPM &  $\pmb{7.00}$  & $\pmb{6.06}$
    \end{tabular} 
    \caption{FID ($\downarrow$) results of StyleGAN2 and DDPMs trained on FFHQ $256^2$ and LSUN Church $256^2$.}
    \label{source_fid}
\end{table}

\begin{figure*}[tbp]
    \centering
    \includegraphics[width=1.0\linewidth]{ }
    \caption{Image samples produced by DDPM-PA models trained for different iterations on 10-shot FFHQ $\rightarrow$ Babies. All the visualized samples of different models are synthesized from fixed noise inputs. }
    \label{amedeo}
\end{figure*}

 In Fig. \ref{scratch2}, we provide samples generated by DDPMs trained from scratch for $60K$ iterations on all three 1000-shot Babies and LSUN Church datasets. In addition, we add the generated samples of models trained from scratch on 1000-shot Babies and LSUN Church, which achieve the best FID results for comparison in Fig. \ref{scratch_compare}. We do not include samples for all three datasets since the model trained for $60K$ iterations on 1000-shot Sunglasses achieves the best FID result as well. For 1000-shot Babies, the model trained for $60K$ iterations achieves smoother results containing fewer blurs despite its worse FID result. The model trained for $30K$ iterations achieves the best FID result and synthesizes more diverse images. As for 1000-shot LSUN Church, the model trained for $50K$ iterations with the best FID result produces samples containing more detailed structures of churches than the model trained for $60K$ iterations. However, all these results are still coarse and lack high-frequency details, indicating the necessity of adapting source models to target domains when training data is limited.

\section{DDPM-based Source Models}
\label{appendix_source}
We train DDPMs on FFHQ $256^2$ \cite{Karras_2020_CVPR} and LSUN Church $256^2$ \cite{yu2015lsun} from scratch for $300K$ iterations and $250K$ iterations as source models for DDPM adaptation, which cost 5 days and 22 hours, 4 days and 22 hours on $\times 8$ NVIDIA RTX A6000 GPUs, respectively.

Image samples produced by these two source models can be found in Fig. \ref{result_scratch}. We randomly sample 1000 images with these two models to evaluate their generation diversity using the average pairwise LPIPS \cite{zhang2018unreasonable} metric, as shown in Table \ref{source_lpips}. For comparison, we also evaluate the generation diversity of the source StyleGAN2 \cite{Karras_2020_CVPR} models used by GAN-based baselines \cite{wang2018transferring, ada, mo2020freeze, wang2020minegan, ewc, ojha2021few-shot-gan, zhao2022closer}. DDPMs trained on FFHQ $256^2$ and LSUN Church $256^2$ achieve generation diversity similar to the widely-used StyleGAN2 models.

Besides, we sample 5000 images to evaluate the generation quality of the source models using FID \cite{heusel2017gans}. As shown in Table \ref{source_fid}, DDPM-based source models achieve FID results similar to StyleGAN2 on the source datasets FFHQ $256^2$ and LSUN Church $256^2$.

 \begin{figure}[t]
    \centering
    \includegraphics[width=1.0\linewidth]{ 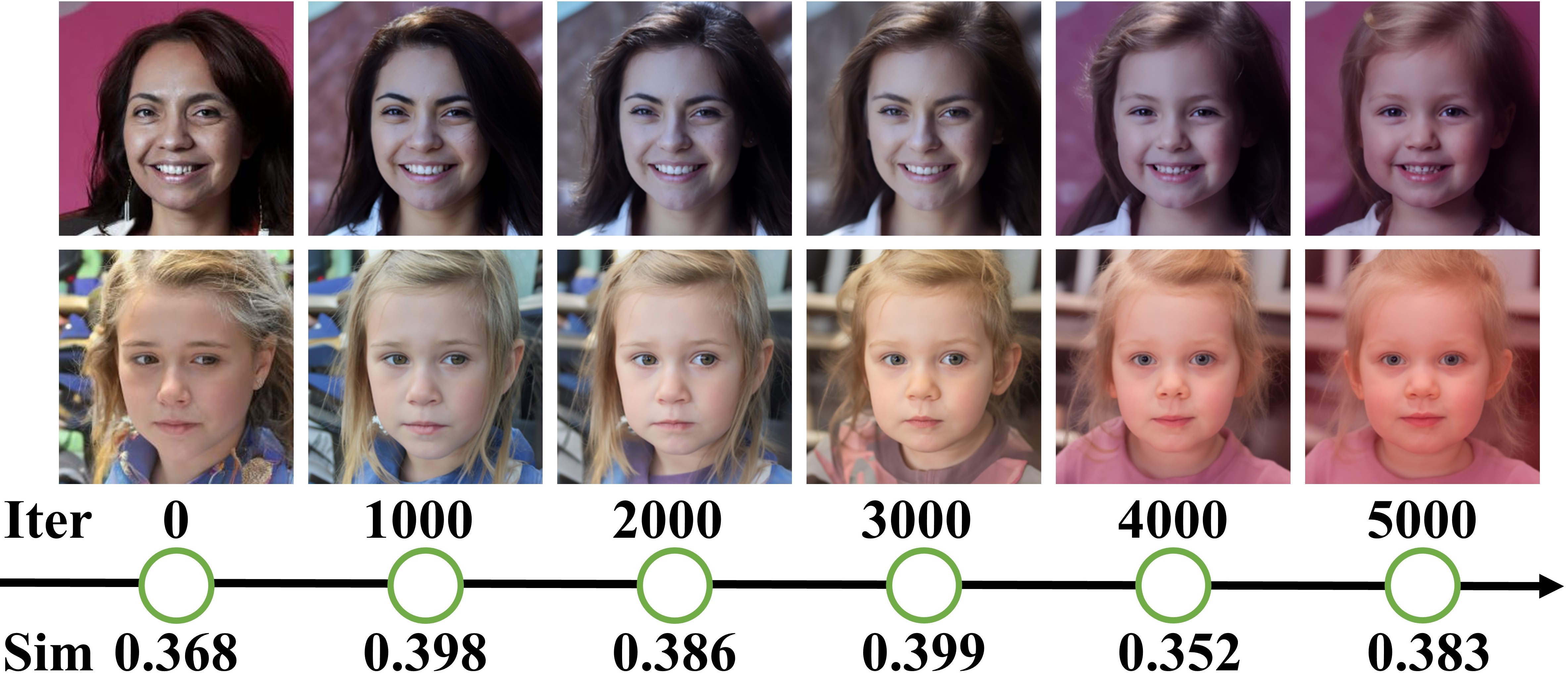}
    \caption{Samples synthesized from fixed noise inputs by DDPM-PA on 10-shot FFHQ $\rightarrow$ Babies. DDPM-PA keeps the relative pairwise distances during domain adaptation and achieves diverse results containing high-frequency details.}
    \label{degrade2}
\end{figure}

 \begin{figure}[t]
    \centering
    \includegraphics[width=1.0\linewidth]{ 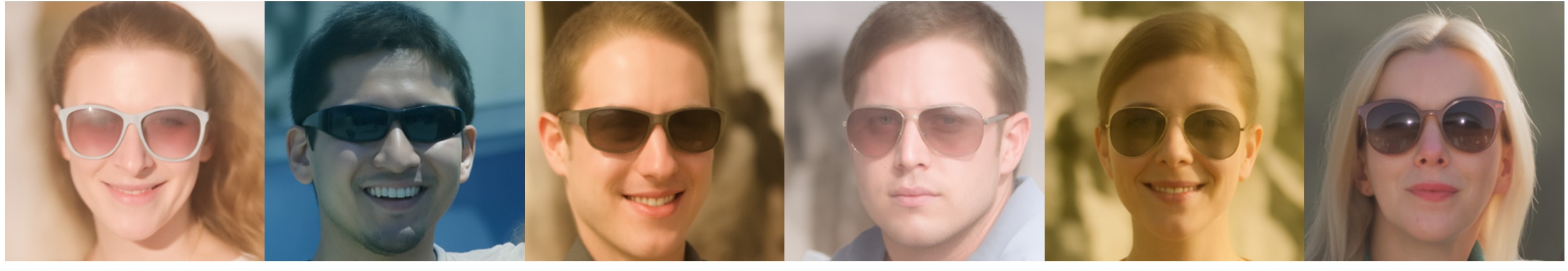}
    \caption{Samples synthesized by DDPM-PA trained on 10-shot FFHQ $\rightarrow$ Sunglasses for 10000 iterations. Too many iterations lead to the replication of training samples and the degradation of generation quality and diversity.}
    \label{sunglass10000}
\end{figure}

\begin{table*}[tbp]
\centering
\begin{tabular}{c|c|c|c|c}
\hline
Datasets & \multicolumn{2}{c|}{1000-shot LSUN Church } &
\multicolumn{2}{c}{1000-shot Sunglasses } \\
\hline
 Iterations (K) & {Scratch} & \makecell[c]{Fine-tuned DDPM \\ (Source Model: FFHQ)} & Scratch & \makecell[c]{Fine-tuned DDPM \\ (Source Model: LSUN Church)} \\
 \hline
 $0$ & $424.53$ & $262.87$ & $419.75$ & $235.21$ \\
 $10$ & $413.68$ & $97.49$ & $419.38$ & $183.86$ \\
 $20$ & $385.25$ & $\pmb{39.68}$ & $348.48$ & $\pmb{56.34}$\\
 $30$ & $388.81$ & $45.74$ & $168.62$ & $60.47$\\
 $40$ & $57.68$ & $51.45$ & $82.48$ & $63.81$\\
 $50$ & $\pmb{45.43}$ & $55.70$ & $68.26$ & $66.45$\\
 $60$ & $69.18$ & $58.43$ & $\pmb{66.09}$ & $66.17$\\
 \hline
\end{tabular}
\caption{FID ($\downarrow$) results of fine-tuned DDPMs trained on unrelated source/target domains compared with DDPMs trained from scratch.}
\label{unrelated}
\end{table*}

\begin{figure*}[tbp]
    \centering
    \includegraphics[width=1.0\linewidth]{ }
    \caption{Image samples synthesized from fixed noise inputs by directly fine-tuned DDPMs trained on 1000-shot LSUN Church $\rightarrow$ Sunglasses for different iterations. }
    \label{church_sunglass}
\end{figure*}

\section{DDPM Adaptation Process Analysis}
\label{appendix_adaptation}
This paper mainly concentrates on the challenging 10-shot image generation tasks. When fine-tuning pre-trained DDPMs on target domains using limited data directly, too many iterations lead to overfitting and seriously degraded diversity. Fine-tuned models trained for about 10K iterations almost exclusively focus on replicating the training samples. Therefore, we train the directly fine-tuned DDPMs for $3K-4K$ iterations to adapt source models to target domains and maintain diversity. However, the directly fine-tuned DDPMs still generate coarse samples lacking details with reasonable iterations.

\begin{figure*}[tbp]
    \centering
    \includegraphics[width=1.0\linewidth]{ }
    \caption{Image samples synthesized from fixed noise inputs by directly fine-tuned DDPMs trained on 1000-shot FFHQ $\rightarrow$ LSUN Church for different iterations. }
    \label{ffhq_church}
\end{figure*}

In Fig. \ref{amedeo}, we provide samples produced by DDPM-PA models trained for different iterations on 10-shot FFHQ $\rightarrow$ Babies. We apply fixed noise inputs to different models for comparison. As the iterations increase, the styles of the generated images become closer to the training samples. Images synthesized from the same noise inputs as Fig. \ref{degrade} are included in red boxes. In addition, detailed evaluation of cosine similarity is added in Fig. \ref{degrade2}. The source samples are adapted to the target domain while keeping relatively stable cosine similarity. Compared with the directly fine-tuned DDPMs, DDPM-PA has a stronger ability to maintain generation diversity and achieve realistic results containing rich details. Nonetheless, too many iterations still lead to the degradation of quality and diversity and details missing, as shown by the samples of the DDPM-PA model trained on 10-shot FFHQ $\rightarrow$ Sunglasses for 10K iterations in Fig. \ref{sunglass10000}. Therefore, we recommend choosing suitable iterations for different adaptation setups (e.g., $4K-5K$ iterations for 10-shot FFHQ $\rightarrow$ Babies) to adapt the pre-trained models to target domains naturally and guarantee the high quality and great diversity of generated samples.

\section{Unrelated Source/Target Domains}
\label{appendix_unrelated}
In this paper, we mainly focus on related source/target domains 
for the few-shot image generation tasks, e.g., FFHQ $\rightarrow$ Sketches and LSUN Church $\rightarrow$ Haunted houses. This section adds experiments using unrelated source/target domains, including FFHQ $\rightarrow$ LSUN Church and LSUN Church $\rightarrow$ Sunglasses. We use target datasets containing 1000 images. Here we only compare directly fine-tuned DDPMs with DDPMs trained from scratch. DDPM-PA is inappropriate for unrelated source/target domains since it is designed to preserve diverse information learned from source domains. FID \cite{zhang2018unreasonable} is employed to evaluate the convergence of models. We aim to find out whether unrelated source models could help fasten the convergence of DDPMs.

 \begin{figure}[t]
    \centering
    \includegraphics[width=1.0\linewidth]{ 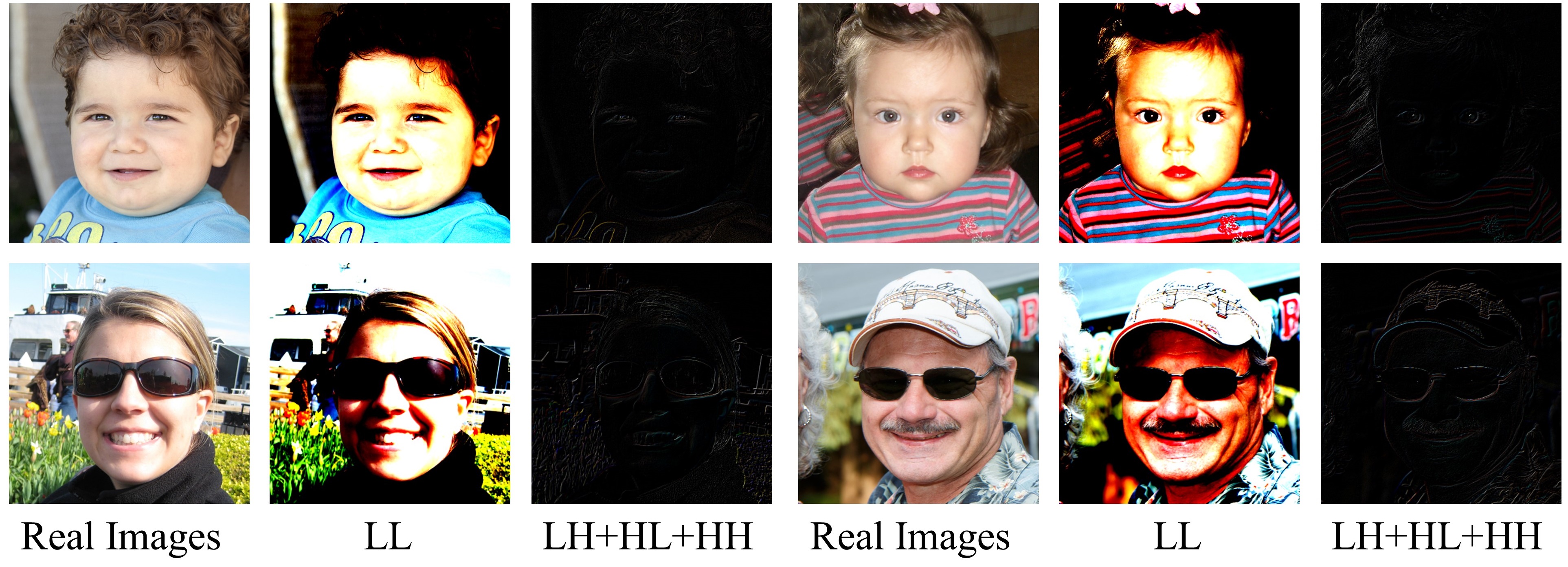}
    \caption{Visualization of the low and high-frequency components obtained with Haar wavelet transformation using images from Babies and Sunglasses as examples. LL represents the low-frequency components, and LH+HL+HH represents the sum of the high-frequency components.}
    \label{wavelet}
\end{figure}

\begin{figure*}[t]
    \centering
    \includegraphics[width=1.0\linewidth]{ 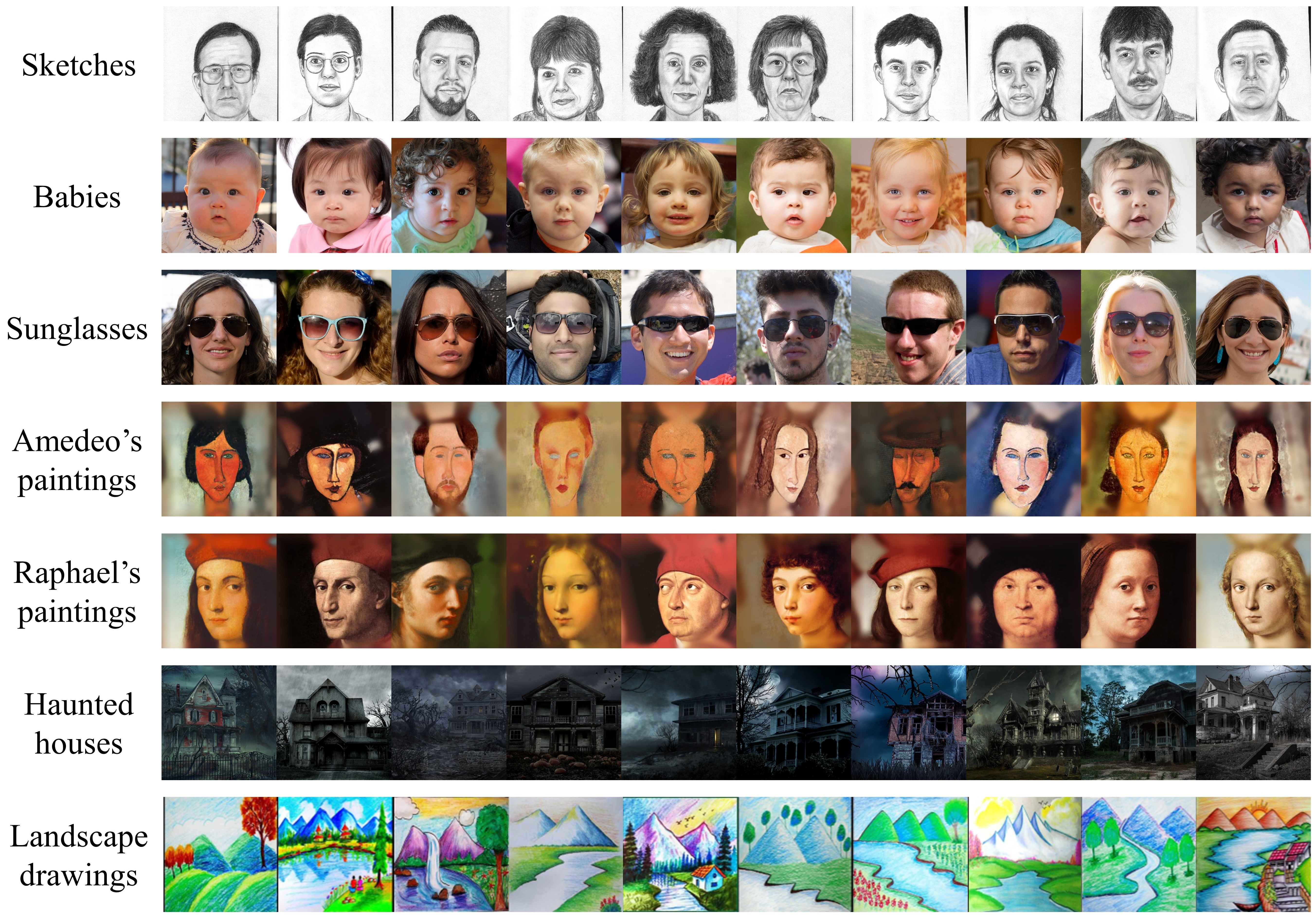}
    \caption{All the 10-shot datasets used in this paper, including 5 target domains corresponding to FFHQ and 2 target domains corresponding to LSUN Church.}
    \label{datasets}
\end{figure*}

As shown in Table \ref{unrelated}, fine-tuned DDPMs using unrelated pre-trained models also accelerate convergence compared with DDPMs trained from scratch. Fine-tuned DDPMs achieve the best FID results using $20K$ iterations for the two employed adaptation setups. The models trained from scratch achieve close FID results when trained for a lot more iterations (about $50K$-$60K$). The adaptation on unrelated source/target domains is still evidently faster than training from scratch. In Fig. \ref{church_sunglass} and \ref{ffhq_church}, we provide visualized samples of fine-tuned DDPMs throughout training and show that DDPMs are capable of adapting unrelated source images to target domains. However, fine-tuned DDPMs still cannot avoid the generation of coarse samples lacking high-frequency details with limited training data in these cases. Besides, too many iterations may lead to blurred results. We recommend DDPM-PA on related source/target domain adaptation to achieve compelling results for few-shot image generation tasks.

\section{Harr Wavelet Transformation Examples}

Fig. \ref{wavelet} visualizes several examples of Haar wavelet transformation. The low-frequency components LL contain the fundamental structures of images. High-frequency components including LH, HL, and HH contain rich details in images.

\section{Computational Cost} 
The time cost of DDPMs and the proposed DDPM-PA approach are listed in Table \ref{timecost}. DDPM-PA costs $24.14\%$ more training time than the original DDPMs. DDPMs trained from scratch need about $40K$ iterations to achieve reasonable results, even if they can only replicate the training samples. DDPM-PA utilizes related pre-trained models to accelerate convergence (about $3K$-$5K$ iterations) and significantly improve generation quality and diversity. Compared with directly fine-tuned DDPMs, DDPM-PA is not overly time-consuming and achieves more realistic results.

\begin{table}[t]
    \centering
    \begin{tabular}{l|c}
        Approaches & Time Cost $/\ 1K$ Iterations   \\
        \hline
        DDPMs  &  29 min   \\
        DDPM-PA &  36 min    
    \end{tabular} 
    \caption{The time cost of directly fine-tuned DDPMs (batch size 48) and DDPM-PA (batch size 24) models trained for $1K$ iterations on $\times 8$ NVIDIA RTX A6000 GPUs.}
    \label{timecost}
\end{table}

\begin{figure*}[t]
    \centering
    \includegraphics[width=1.0\linewidth]{ 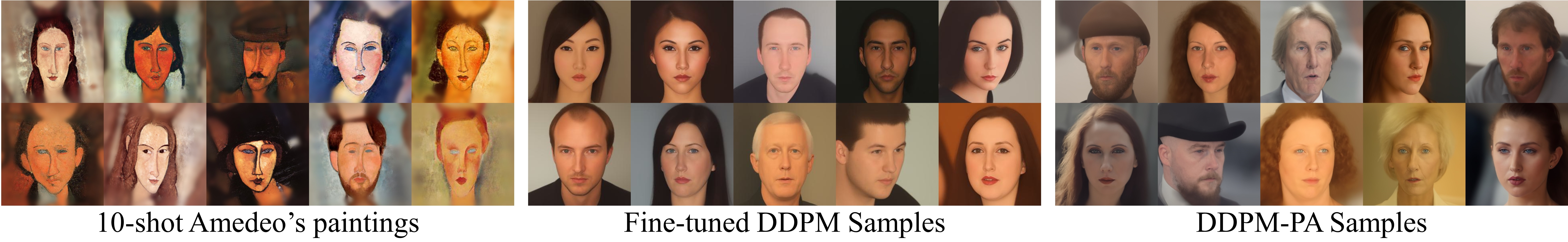}
    \caption{Samples of DDPM-based image generation approaches (directly fine-tuned DDPM and DDPM-PA) on 10-shot FFHQ $\rightarrow$ Amedeo's paintings.}
    \label{sketches}
\end{figure*}

\begin{figure*}[t]
    \centering
    \includegraphics[width=1.0\linewidth]{ 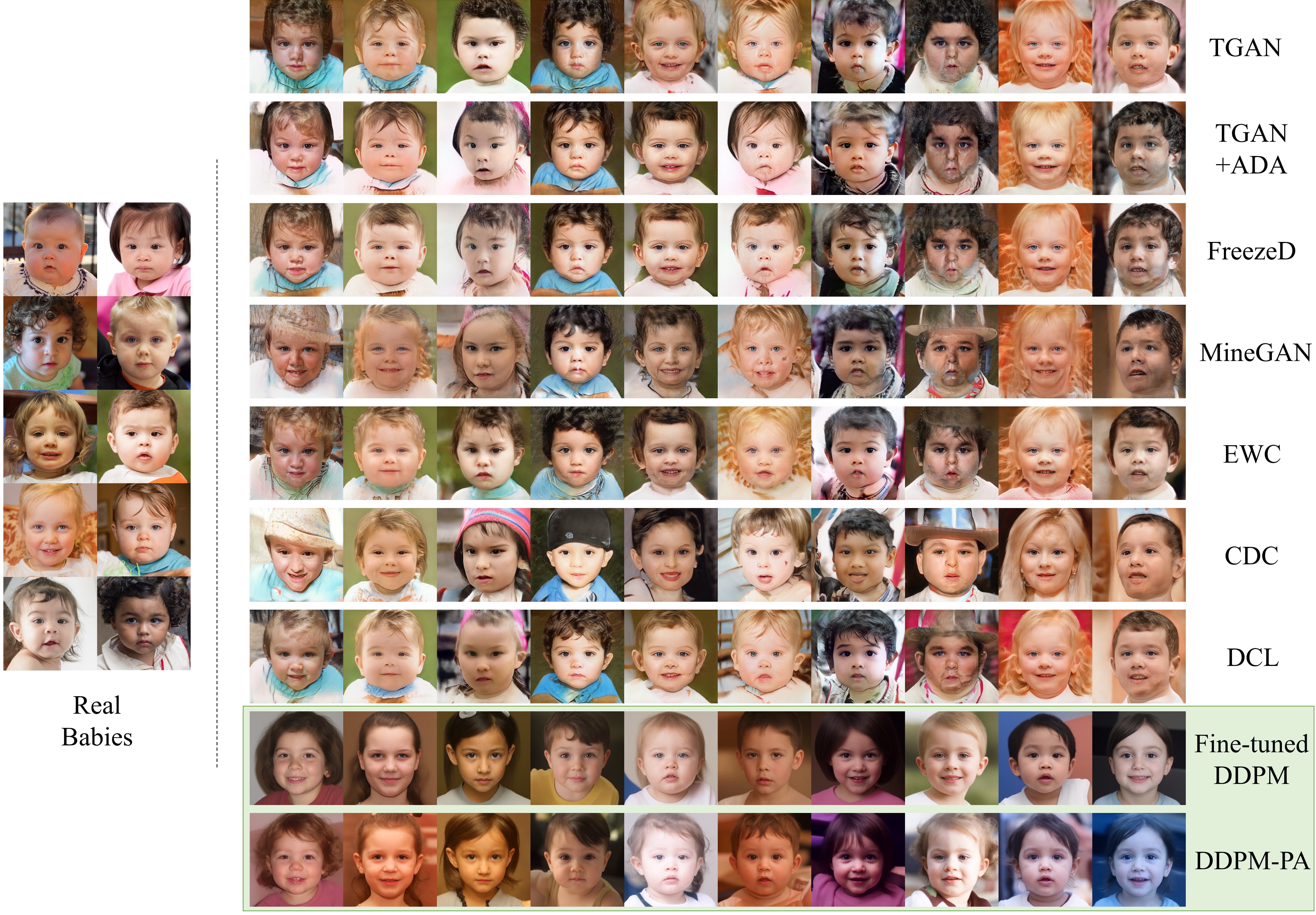}
    \caption{10-shot image generation samples on FFHQ $\rightarrow$ Babies. All the samples of GAN-based approaches are synthesized from fixed noise inputs (rows 1-7). Samples of the directly fine-tuned DDPM and DDPM-PA are synthesized from fixed noise inputs as well (rows 8-9). }
    \label{babies}
\end{figure*}

\section{Additional Visualized Samples }
\label{appendix_results}
We show all the 10-shot datasets used in this paper for few-shot image generation tasks in Fig. \ref{datasets}, including 5 target domains corresponding to the source domain FFHQ \cite{Karras_2020_CVPR} and 2 target domains corresponding to LSUN Church \cite{yu2015lsun}.

In Fig. \ref{sketches}, we provide image generation samples of DDPM-PA on 10-shot FFHQ $\rightarrow$ Amedeo's paintings as supplements to Fig. \ref{result1}. Compared with directly fine-tuned DDPMs, DDPM-PA generates more diverse samples containing richer details. DDPM-PA encourages adapted models to learn the common features of limited training samples while maintaining considerable generation diversity.

\begin{figure*}[t]
    \centering
    \includegraphics[width=1.0\linewidth]{ 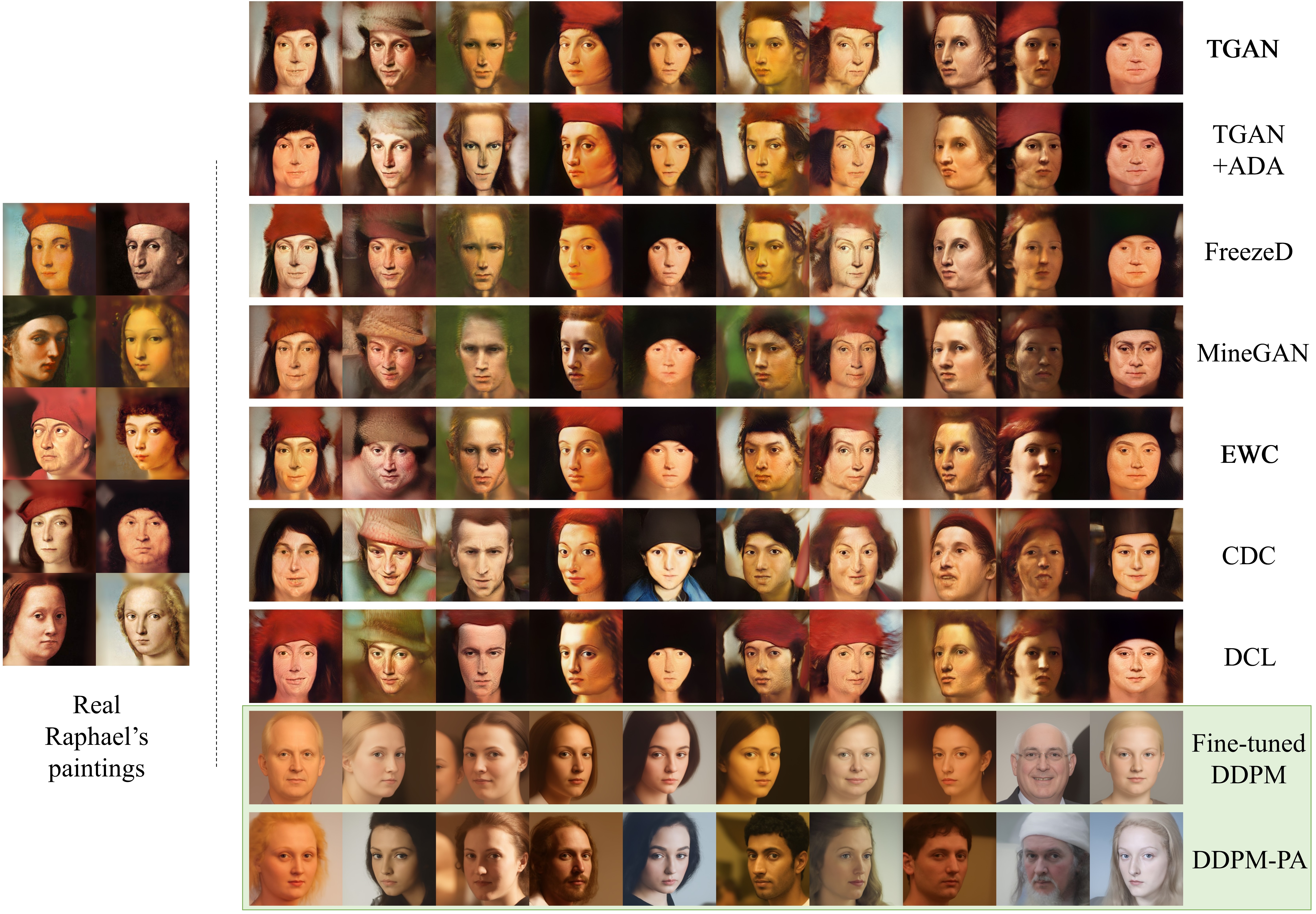}
    \caption{10-shot image generation samples on FFHQ $\rightarrow$ Raphael's paintings. All the samples of GAN-based approaches are synthesized from fixed noise inputs (rows 1-7). Samples of the directly fine-tuned DDPM and DDPM-PA are synthesized from fixed noise inputs as well (rows 8-9). }
    \label{raphael}
\end{figure*}

\begin{figure*}[t]
    \centering
    \includegraphics[width=1.0\linewidth]{ 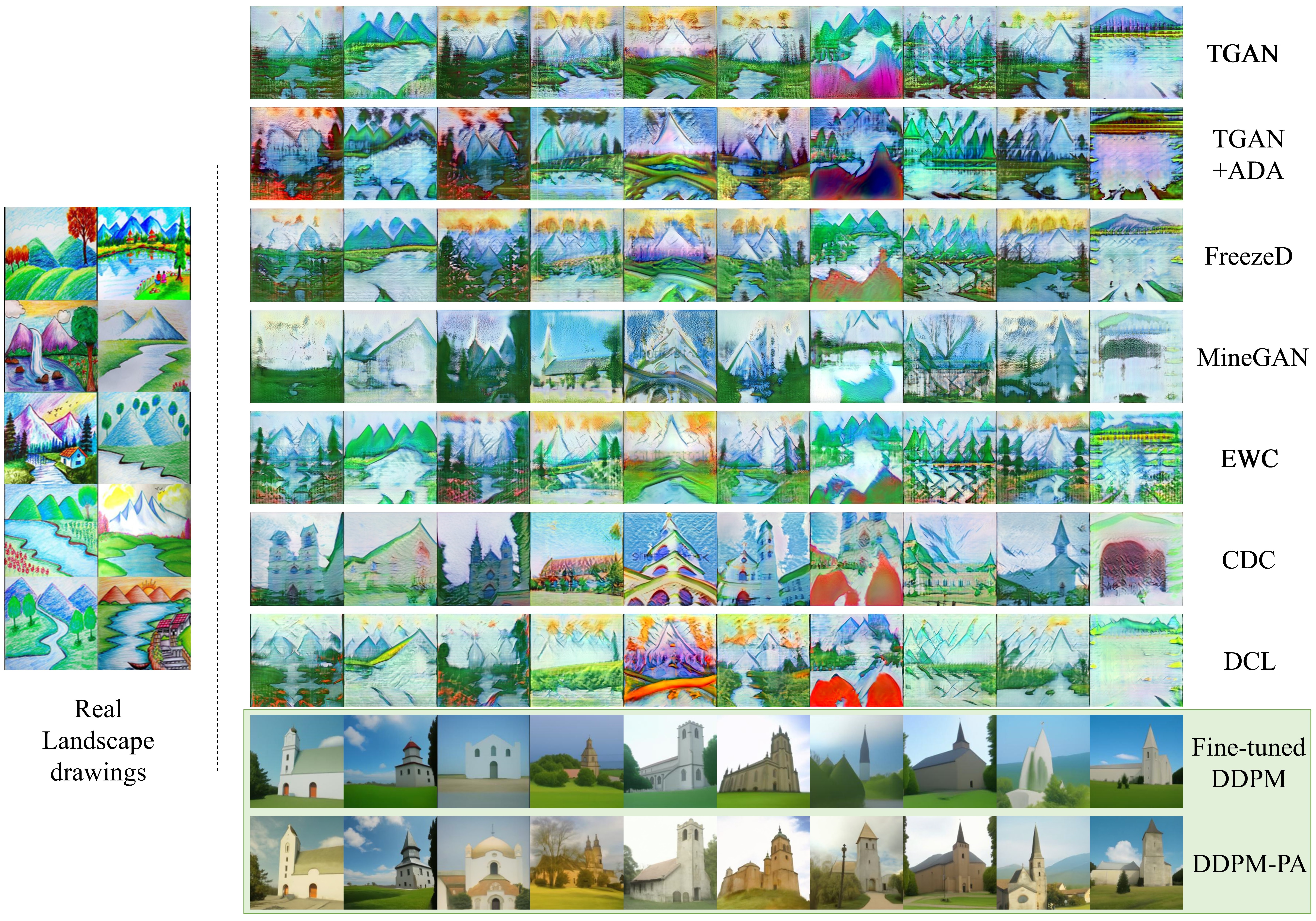}
    \caption{10-shot image generation samples on LSUN Church $\rightarrow$ Landscape drawings. All the samples of GAN-based approaches are synthesized from fixed noise inputs (rows 1-7). Samples of the directly fine-tuned DDPM and DDPM-PA are synthesized from fixed noise inputs as well (rows 8-9). }
    \label{vangogh}
\end{figure*}

Besides, we provide image generation samples of GAN-based baselines and DDPM-based approaches on 10-shot FFHQ $\rightarrow$ Babies, FFHQ $\rightarrow$ Raphael's paintings, and LSUN Church $\rightarrow$ Landscape drawings in Fig. \ref{babies}, \ref{raphael}, and \ref{vangogh} as supplements to Fig. \ref{sunglass}. We apply fixed noise inputs to GAN-based approaches and DDPM-based approaches, respectively. DDPMs are more stable and less vulnerable to overfitting than GANs. GANs easily overfit and tend to generate samples similar to training samples when training data is limited (see samples of TGAN \cite{wang2018transferring}). Directly fine-tuned DDPMs can still keep a measure of generation diversity under the same conditions. Besides,  DDPM-based approaches avoid the generation of blurs and artifacts. However, directly fine-tuned DDPMs tend to produce too smooth results lacking high-frequency details and still face diversity degradation. DDPM-PA generates more realistic results containing richer high-frequency details than GAN-based baselines under all these adaptation setups.

DDPM-PA performs apparently better when we want to keep the fundamental structures of source images and learn the styles of target domains (e.g., LSUN Church $\rightarrow$ Landscape drawings). As shown in Fig. \ref{vangogh}, GAN-based approaches fail to adapt real churches to the drawing style and produce samples containing too many blurs and artifacts. On the other hand, DDPM-PA produces high-quality church drawings and preserves more detailed building structures.

\end{document}